\documentclass[10pt,twocolumn,letterpaper]{article}

\usepackage{cvpr}
\usepackage{times}
\usepackage{epsfig}
\usepackage{graphicx}
\usepackage{amsmath}
\usepackage{subfig}
\usepackage{textcomp}
\usepackage{amssymb}
\usepackage{breqn}
\usepackage[noline,boxed]{algorithm2e}
\usepackage[toc,page]{appendix}
\usepackage{balance}
\usepackage{arydshln}

\DeclareMathOperator{\vx}{\mathbf{x}}
\DeclareMathOperator{\vz}{\mathbf{z}}
\DeclareMathOperator{\vy}{\mathbf{y}}

\DeclareMathOperator{\vmu}{\boldsymbol{\mu}}

\DeclareMathOperator{\R}{\mathbb{R}}
\DeclareMathOperator{\N}{\mathcal{N}}

\usepackage[pagebackref=true,breaklinks=true,letterpaper=true,colorlinks,bookmarks=false]{hyperref}

\cvprfinalcopy 

\def\cvprPaperID{10195} 

\ifcvprfinal\fi
\begin{document}


\title{Improved Few-Shot Visual Classification}

\author{Peyman Bateni\textsuperscript{1},
Raghav Goyal\textsuperscript{1,3}, Vaden Masrani\textsuperscript{1}, Frank Wood\textsuperscript{1,2,4}, Leonid Sigal\textsuperscript{1,3,4}\\
\textsuperscript{1}University of British Columbia, \textsuperscript{2}MILA, \textsuperscript{3}Vector Institute, \textsuperscript{4}CIFAR AI Chair\\
{\tt\small \{pbateni, rgoyal14, vadmas, fwood, lsigal\}@cs.ubc.ca}
}

\maketitle

\begin{abstract}
   Few-shot learning is a fundamental task in computer vision that carries the promise of alleviating the need for exhaustively labeled data.  Most few-shot learning approaches to date have focused on progressively more complex neural feature extractors and classifier adaptation strategies, and the refinement of the task definition itself. In this paper, we explore the hypothesis that a simple class-covariance-based distance metric, namely the Mahalanobis distance, adopted into a state of the art few-shot learning approach (CNAPS \cite{requeima2019fast}) can, in and of itself, lead to a significant performance improvement.  We also discover that it is possible to learn adaptive feature extractors that allow useful estimation of the high dimensional feature covariances required by this metric from surprisingly few samples. The result of our work is a new ``Simple CNAPS'' architecture which has up to 9.2\% fewer trainable parameters than CNAPS and performs up to 6.1\% better than state of the art on the standard few-shot image classification benchmark dataset.
\end{abstract}

\section{Introduction}

Deep learning successes have led to major computer vision advances \cite{Hossain:2019:CSD:3303862.3295748-image-captioning-survey,DBLP:journals/corr/abs-1907-09408-object-detection-survey,8441512-image-classification-survey}. However, most methods behind these successes have to operate in fully-supervised, high data availability regimes. This limits the applicability of these methods, effectively excluding domains where data is fundamentally scarce or impossible to label en masse.  This inspired the field of few-shot learning  \cite{Wang:2019:SZL:3306498.3293318-survey-of-zero-shot-learning, DBLP:journals/corr/abs-1904-05046-survey-on-few-shot-learning} which aims to computationally mimic  human reasoning and learning from limited data. 

The goal of few-shot learning is to automatically adapt models such that they work well on instances from classes not seen at training time, given only a few labelled examples for each new class. In this paper, we focus on few-shot image classification where the ultimate aim is to develop a classification methodology that automatically adapts to new classification tasks at test time, and particularly in the case where only a very small number of labelled ``support'' images are available per class.

\begin{figure}%
    \centering
    \subfloat[Squared Euclidean Distance]{{\includegraphics[width=4.0cm]{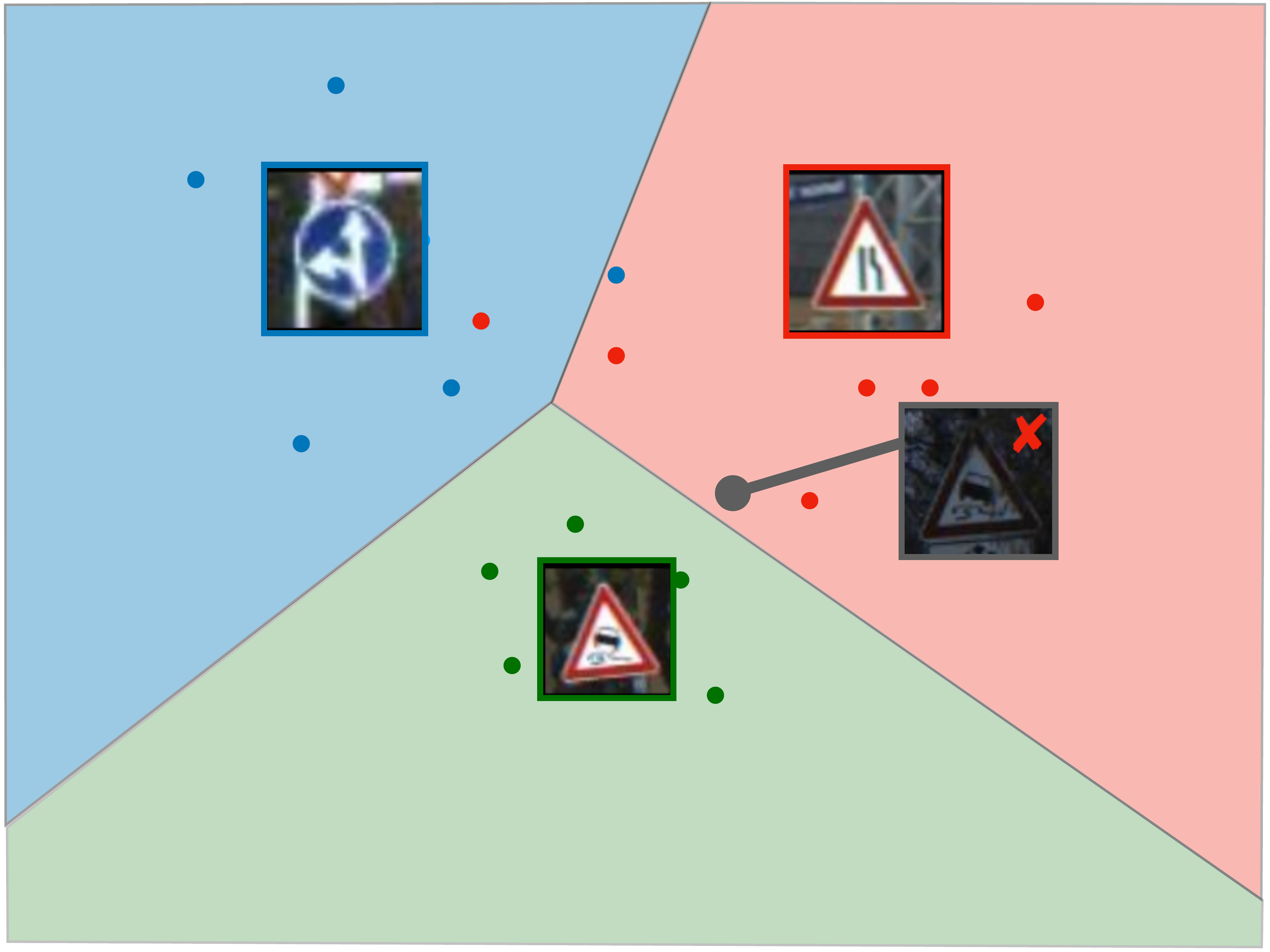} }}%
    \subfloat[Squared Mahalanobis Distance]{{\includegraphics[width=4.0cm]{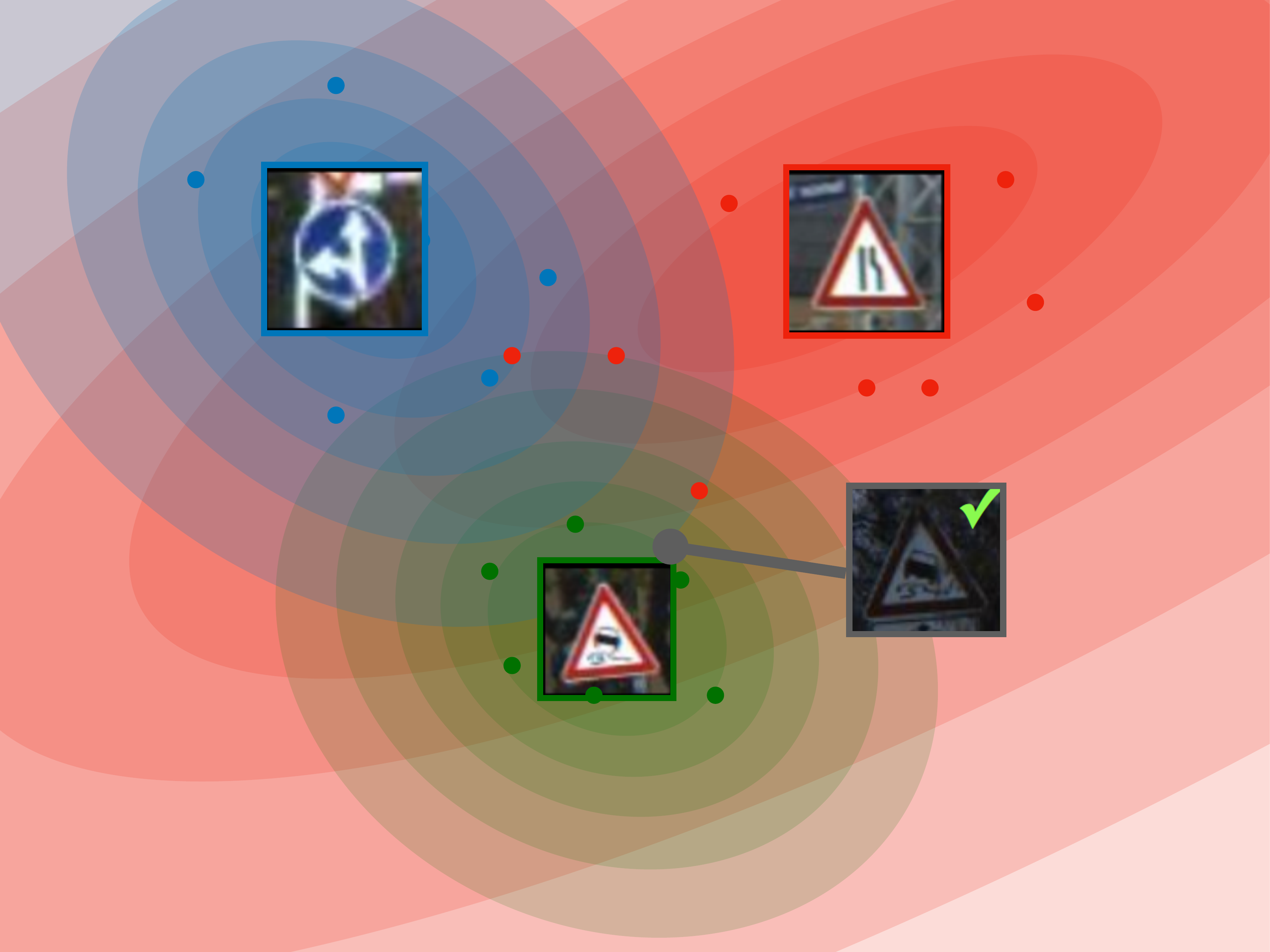} }}%
    \vspace{-0.1in}
    \caption{{\bf Class-covariance metric:} Two-dimensional illustration of the embedded support image features  output by a task-adapted feature extractor (points),  per-class embedding means (inset icons), explicit (left) and implied class decision boundaries (right), and test query instance (gray point and inset icon) for two  classifiers: standard $L_2^2$-based (left) and ours, class-covariance-based (Mahalanobis distance, right).  An advantage of using a class-covariance-based  metric during classification is that taking into account the distribution in feature space of each class can result in improved non-linear classifier decision boundaries.  What cannot explicitly appear in this figure, but we wish to convey here regardless, is that the task-adaptation mechanism used to produce these embeddings is trained end-to-end from the Mahalanobis-distance-based classification loss. This means that, in effect, the task-adaptation feature extraction mechanism learns to produce embeddings that result in informative task-adapted covariance estimates.}
    \vspace{-0.1in}
\end{figure}

\begin{figure}[t]
    \centering
    \includegraphics[width=3.2in]{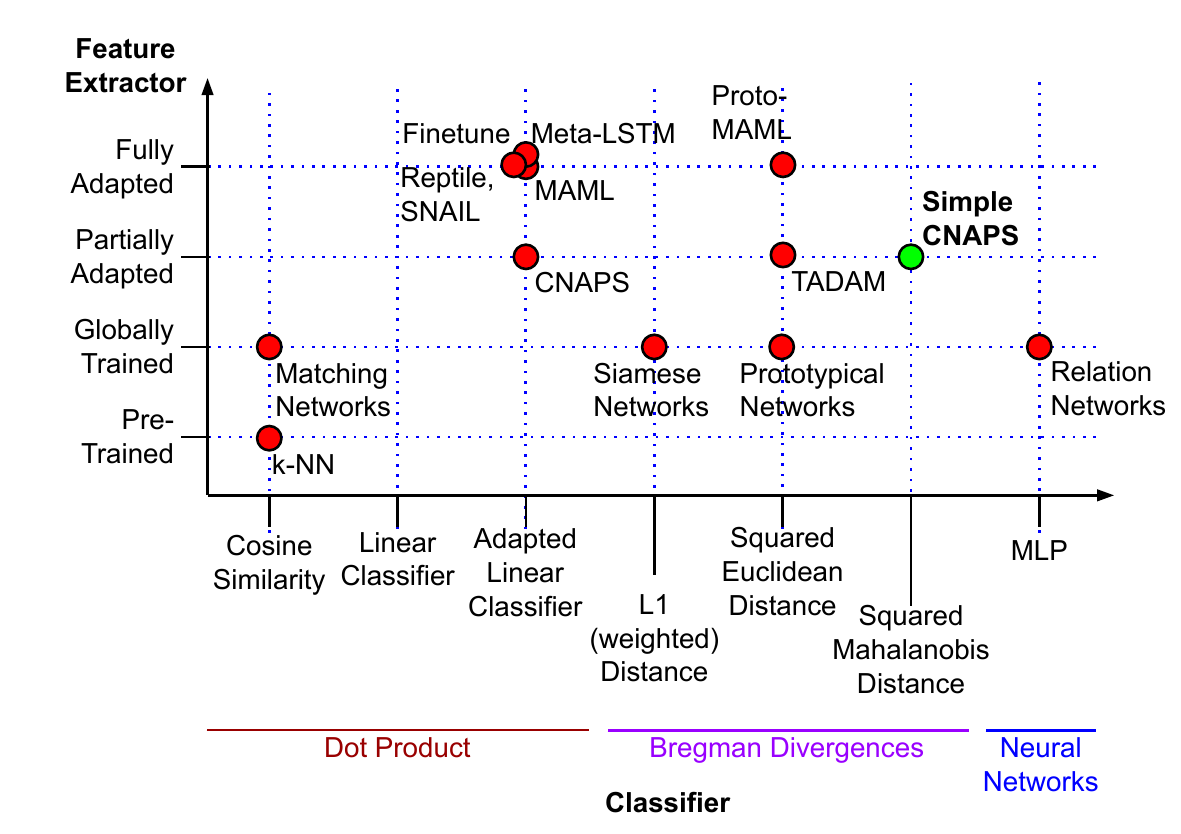}
    \vspace{-0.1in}
    \caption{{\bf Approaches to few-shot image classification:} organized by image feature extractor adaptation scheme (vertical axis) versus final classification method (horizontal axis). Our method (Simple CNAPS) partially adapts the feature extractor (which is architecturally identical to CNAPS) but is trained with, and uses, a fixed, rather than adapted, Mahalanobis metric for final classification.}
    \vspace{-0.1in}
    \label{related-work-overview}
\end{figure}

Few-shot learning approaches typically take one of two forms: 1) nearest neighbor approaches and their variants, including matching networks \cite{vinyals2016matching}, which effectively apply nearest-neighbor or weighted nearest neighbor classification on the samples themselves, either in a feature \cite{DBLP:journals/corr/abs-1905-01436-edge-labelling-gnn,koch2015siamese, garcia2018fewshot} or a semantic space \cite{Frome-NIPS2013_5204}; or 2) embedding methods that effectively distill all of the examples to a single prototype per class, where a prototype may be learned \cite{gidaris2019generating, requeima2019fast} or implicitly derived from the samples \cite{snell2017prototypical} (\eg~mean embedding). The prototypes are often defined in feature or semantic space (\eg word2vec \cite{DBLP:journals/corr/abs-1902-07104-elementai}). Most research in this domain has focused on  learning non-linear mappings, often expressed as neural nets, from images to the embedding space subject to a pre-defined metric in the embedding space used for final nearest class classification; usually cosine similarity between query image embedding and class embedding. Most recently, CNAPS \cite{requeima2019fast} achieved state of the art (SoTA) few-shot visual image classification by  utilizing sparse FiLM \cite{perez2018film} layers within the context of episodic training to avoid problems that arise from trying to adapt the entire embedding network using few support samples.

Overall much less attention has been given to the metric used to compute distances for classification in the embedding space. Presumably this is because common wisdom dictates that flexible non-linear mappings are ostensibly able to adapt to any such metric, making the choice of metric apparently inconsequential. In practice, as we find in this paper, the choice of metric is quite important. In \cite{snell2017prototypical} the authors analyze the underlying distance function used in order to justify the use of sample means as prototypes. They argue that Bregman divergences \cite{banerjee2005clustering} are the theoretically sound family of metrics to use in this setting, but only utilize a single instance within this class — squared Euclidean distance, which they find to perform better than the more traditional cosine metric. However, the choice of Euclidean metric involves making two flawed assumptions: 1) that feature dimensions are un-correlated and 2) that they have uniform variance. Also, it is insensitive to the distribution of within-class samples with respect to their prototype and recent results \cite{NIPS2018_7352-tadam,snell2017prototypical} suggest that this is problematic.   Modeling this distribution (in the case of \cite{banerjee2005clustering} using extreme value theory) is, as we find, a key to better performance.

\vspace{0.05in}
\noindent
{\bf Our Contributions:}
Our contributions are four-fold: 
1) A robust empirical finding of a significant 6.1\%  improvement, on average, over SoTA (CNAPS \cite{requeima2019fast}) in few-shot image classification, obtained by utilizing a test-time-estimated class-covariance-based distance metric, namely the Mahalanobis distance \cite{galeano2015mahalanobis}, in final, task-adapted classification. 2) The surprising finding that we are able to estimate such a metric even in the few shot classification setting, where the number of available support examples, per class, is far too few in theory to estimate the required class-specific covariances. 3) A new ``Simple CNAPS'' architecture that achieves this performance despite removing 788,485 parameters (3.2\%-9.2\% of the total) from original CNAPS architecture, replacing them with fixed, not-learned, deterministic covariance estimation and Mahalanobis distance computations. 4) Evidence that should make readers question the common understanding that CNN feature extractors of sufficient complexity can adapt to any final metric (be it cosine similarity/dot product or otherwise).

\section{Related Work}
\label{sec:related-work}

\begin{figure*}
    \centering
    \includegraphics[width=6.6in]{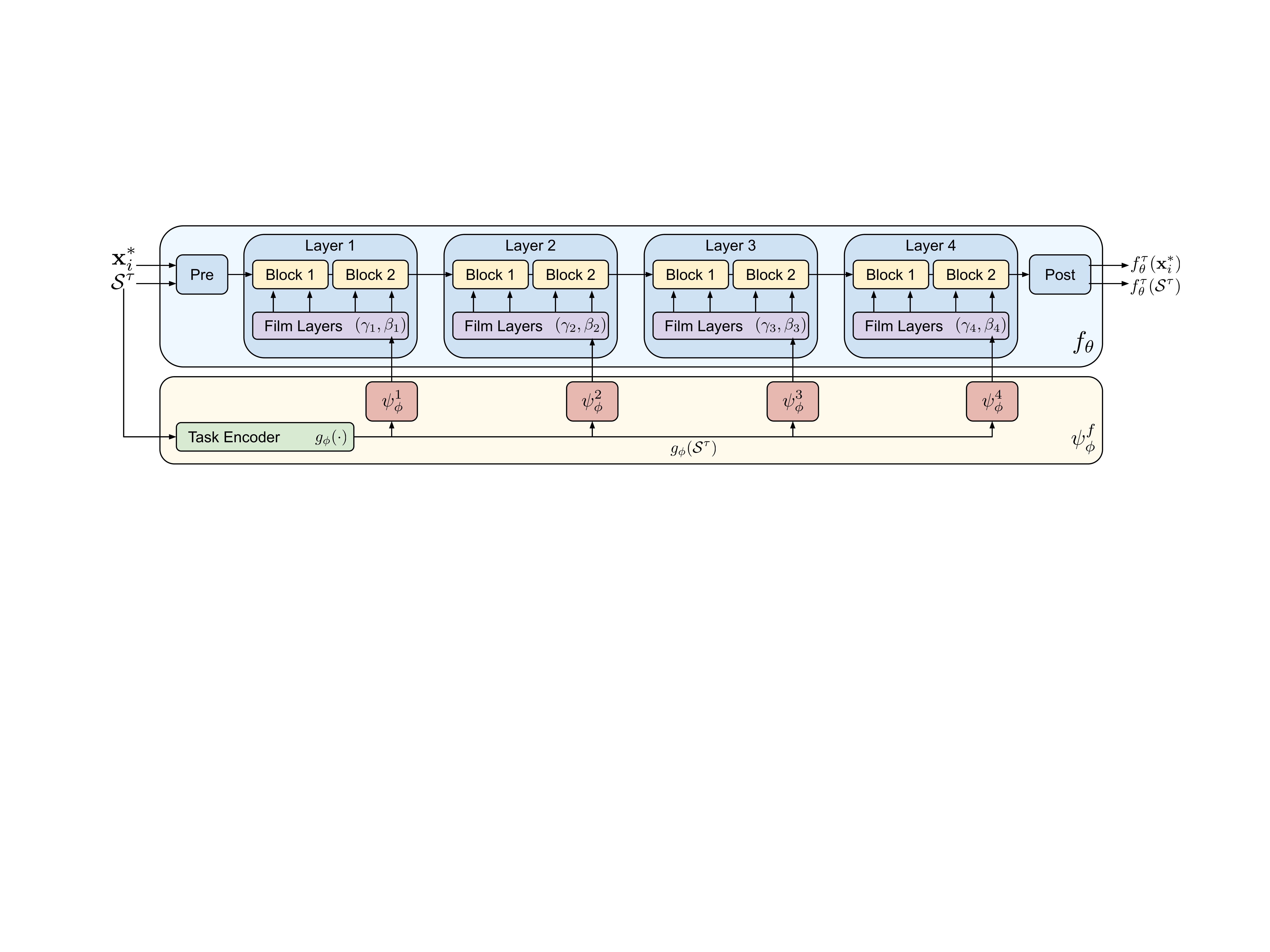}
    \vspace{-0.1in}
    \caption{{\bf Overview of the feature extractor adaptation methodology in CNAPS:} task encoder $g_\phi(\cdot)$ provides the adaptation network $\psi_\phi^i$ at each block $i$ with the task representations $(g_\phi(\mathcal{S}^\tau)$ to produce FiLM parameters $({\gamma}_j, {\beta}_j)$. For details on the auto-regressive variant (AR-CNAPS), architectural implementations, and FiLM layers see Appendix \ref{appendix:cnaps-details}. For an in-depth explanation, refer to the original paper \cite{requeima2019fast}.}
    \label{cnaps-feature-extraction-overview}
    \vspace{-0.1in}
\end{figure*}

Most of last decade's few-shot learning works \cite{DBLP:journals/corr/abs-1904-05046-survey-on-few-shot-learning} can be differentiated along two main axes: 1) how images are transformed into vectorized embeddings, and 2) how ``distances'' are computed between vectors in order to assign labels. This is shown in Figure~\ref{related-work-overview}.

Siamese networks \cite{koch2015siamese}, an early approach to few-shot learning and classification,  used a shared feature extractor to produce embeddings for both the support and query images. Classification was then done by picking the smallest weighted L1 distance between query and labelled image embeddings.
Relation networks \cite{sung2018learning}, and recent GCNN variants \cite{DBLP:journals/corr/abs-1905-01436-edge-labelling-gnn, garcia2018fewshot}, extended this by parameterizing and learning the classification metric using a Multi-Layer Perceptron (MLP).  Matching networks \cite{vinyals2016matching} learned distinct feature extractors for support and query images which were then used to compute cosine similarities for classification.

The feature extractors used by these models were, notably, not adapted to test-time classification tasks. It has become established that adapting feature extraction to new tasks at test time is generally a good thing to do.  Fine tuning transfer-learned networks \cite{DBLP:journals/corr/YosinskiCBL14-finetune} did this by fine-tuning the feature extractor network using the task-specific support images but found limited success due to problems related to overfitting to, the generally very few, support examples. MAML \cite{finn2017model} (and its many extensions \cite{ DBLP:journals/corr/MishraRCA17-snail, DBLP:journals/corr/abs-1803-02999-reptile,DBLP:conf/iclr/RaviL17-meta-lstm}) mitigated this issue by learning a set of meta-parameters that specifically enabled the feature extractors to be adapted to new tasks given few support examples using few gradient steps.

The two methods most similar to our own are CNAPS \cite{requeima2019fast} (and the related TADAM \cite{NIPS2018_7352-tadam}) and Prototypical networks~\cite{snell2017prototypical}. CNAPS is a few-shot adaptive classifier based on conditional neural processes (CNP) \cite{DBLP:journals/corr/abs-1807-01613-cnp}.  It is the state of the art approach for few-shot image classification \cite{requeima2019fast}.  It uses a pre-trained feature extractor augmented with  FiLM layers \cite{perez2018film} that are adapted for each task using the support images specific to that task.  CNAPS uses a dot-product distance in a final linear classifier; the parameters of which are also adapted at test-time to each new task. We describe CNAPS in greater detail when describing our method. 

Prototypical networks \cite{snell2017prototypical} do not use a feature adaptation network; they instead use a simple mean pool operation to form class ``prototypes.''  Squared Euclidean distances to these prototypes are then subsequently used for classification. Their choice of the distance metric was motivated by the theoretical properties of Bregman divergences \cite{banerjee2005clustering}, a family of functions of which the squared Euclidean distance is a member of. These properties allow for a mathematical correspondence between the use of the squared Euclidean distance in a Softmax classifier and performing density estimation. Expanding on \cite{snell2017prototypical} in our paper, we also exploit similar properties of the squared Mahalanobis distance as a Bregman divergence \cite{banerjee2005clustering} to draw theoretical connections to multi-variate Gaussian mixture models.

\begin{figure*}[t]
    \centering
    \includegraphics[width=6.5in]{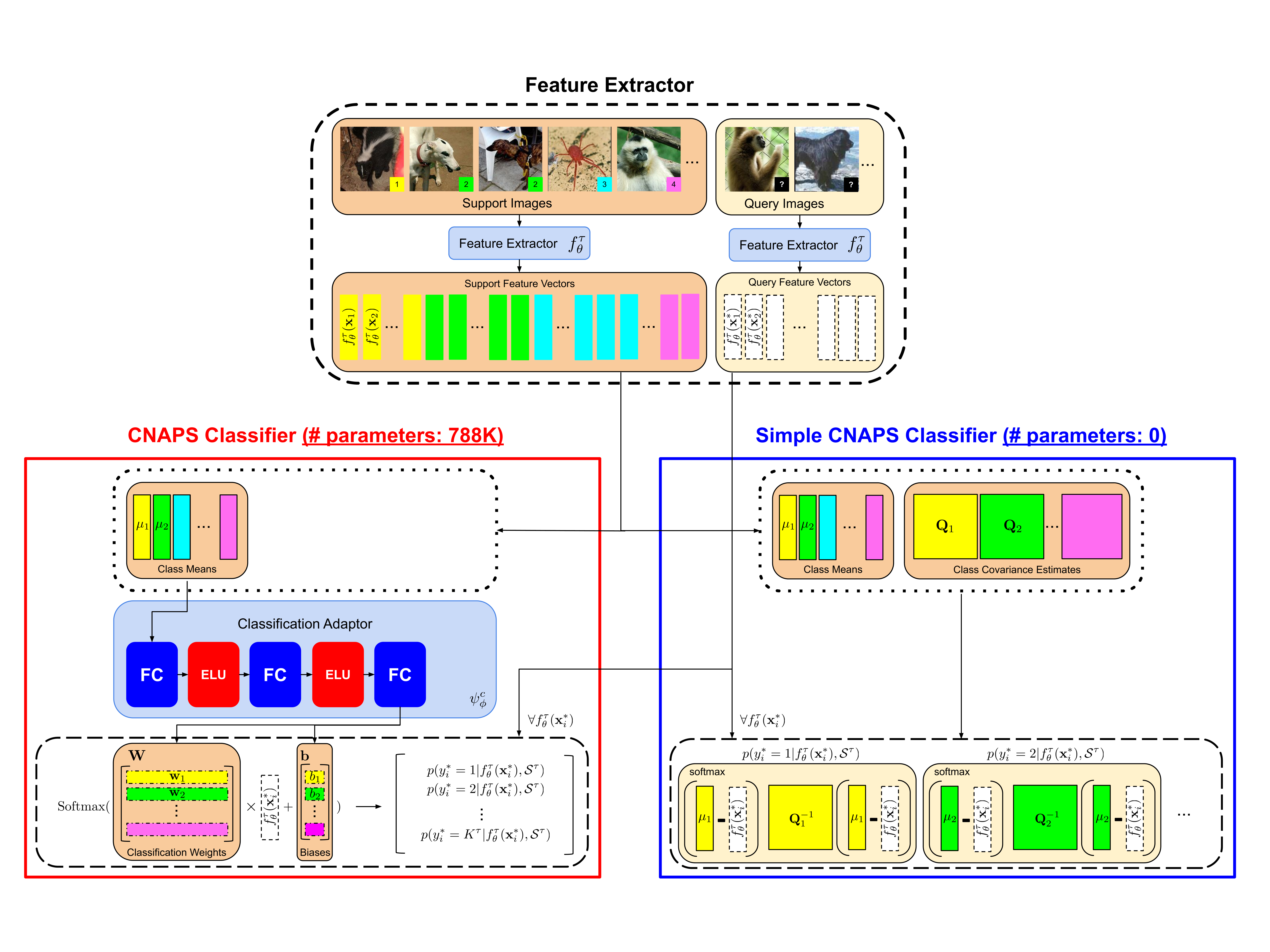}
    \vspace{-0.1in}
    \caption{{\bf Comparison of the feature extraction and classification in CNAPS versus Simple CNAPS:} Both CNAPS and Simple CNAPS share the feature extraction adaptation architecture detailed in Figure~\ref{cnaps-feature-extraction-overview}. CNAPS and Simple CNAPS differ in how distances between query feature vectors and class feature representations are computed for classification.  CNAPS uses a trained, adapted linear classifier whereas Simple CNAPS uses a differentiable but fixed and parameter-free deterministic distance computation.  Components in light blue have parameters that are trained, specifically $f_\theta^\tau$ in both models and $\psi_\phi^c$ in the CNAPS adaptive classification. CNAPS classification requires $778$k parameters while Simple CNAPS is fully deterministic.}
    \label{cnaps-vs-us-direct-comparison}
    \vspace{-0.135in}
\end{figure*}

Our work differs from CNAPS \cite{requeima2019fast} and Prototypical networks \cite{snell2017prototypical} in the following ways. First, while CNAPS has demonstrated the importance of adapting the feature extractor to a specific task, we show that adapting the classifier is actually unnecessary to obtain good performance. Second, we demonstrate that an improved choice of Bregman divergence can significantly impact accuracy. Specifically we show that regularized class-specific covariance estimation from task-specific adapted feature vectors allows the use of the Mahalanobis distance for classification, achieving a significant improvement over state of the art. A high-level diagrammatic comparison of our ``Simple CNAPS'' architecture to CNAPS can be found in Figure~\ref{cnaps-vs-us-direct-comparison}. 

More recently, \cite{DBLP:journals/corr/abs-1708-02735-fort} also explored using the Mahalanobis distance by incorporating its use in Prototypical networks \cite{snell2017prototypical}. In particular they used a neural network to produce per-class diagonal covariance estimates, however, this approach is restrictive and limits performance. Unlike \cite{DBLP:journals/corr/abs-1708-02735-fort}, Simple CNAPS generates regularized full covariance estimates from an end-to-end trained adaptation network.

{}

\section{Formal Problem Definition}
\label{sec:problem_definition}
We frame few-shot image classification as an amortized classification task.  Assume that we have a large labelled dataset $\mathcal{D} = \{ (\vx_i, y_i )\}_{i=1}^{N}$ of images $\vx_i$ and labels $y_i$.  From this dataset we can construct a very large number of classification {\em tasks} $\mathcal{D}^{\tau}  \subseteq \mathcal{D}$ by repeatedly sampling without replacement from $\mathcal{D}$.  Let $\tau\in\mathbb{Z}_+$ uniquely identify a classification task.  We define the {\em support set} of a task to be $\mathcal{S}^{\tau} = \{ (\vx_i, y_i )\}_{i=1}^{N^\tau}$ and the \textit{query set} $Q^{\tau} = \{ (\vx^*_i, y^*_i) \}_{i=1}^{N^{*\tau}}$ where  $\mathcal{D}^{\tau} = \mathcal{S}^{\tau} \cup Q^{\tau}$ where $\vx_i,\vx^*_i \in \R^D$ are vectorized images and $y_i, y_i^* \in \{1,...,K\}$ are class labels.  Our objective is to find parameters $\theta$ of a classifier $f_\theta$ that maximizes
$\mathbb{E}_\tau[\prod_{Q^\tau} p(y^*_i| f_\theta(\vx^*_i, \mathcal{S}^{\tau} )]$.

In practice, $\mathcal{D}$ is constructed by concatenating large image classification datasets and the set of classification tasks. $\{\mathcal{D}^\tau\}_{\tau=1}$ is sampled in a more complex way than simply without replacement.  In particular, constraints are placed on the relationship of the image label pairs present in the support set and those present in the query set.  For instance, in few-shot learning, the constraint that the query set labels are a subset of the support set labels is imposed. %
With this constraint imposed, the classification task reduces to correctly assigning each query set image to one of the classes present in the support set.  Also, in this constrained few-shot classification case, the support set can be interpreted as being the ``training data'' for implicitly training (or adapting) a task-specific classifier of query set images. Note that in conjecture with \cite{requeima2019fast, triantafillou2019meta} and unlike earlier work \cite{snell2017prototypical, finn2017model, vinyals2016matching}, we do not impose any constraints on the support set having to be balanced and of uniform number of classes, although we do conduct experiments on this narrower setting too.

\section{Method}
\label{sec:method}
Our classifier shares feature adaptation architecture with CNAPS \cite{requeima2019fast}, but deviates from CNAPS by replacing their adaptive classifier with a simpler classification scheme based on estimating Mahalanobis distances. To explain our classifier, namely ``Simple CNAPS'', we first detail CNAPS in Section~\ref{sec:cnaps}, before presenting our model in Section~\ref{sec:simple-cnaps}.

\subsection{CNAPS}
\label{sec:cnaps}
Conditional Neural Adapative Processes (CNAPS) consist of two elements: a feature extractor and a classifier, both of which are task-adapted.  Adaptation is performed by trained adaptation modules that take the support set. 

The feature extractor architecture used in both CNAPS and Simple CNAPS is shown in Figure \ref{cnaps-feature-extraction-overview}. It consists of a ResNet18 \cite{DBLP:journals/corr/HeZRS15-resnet} network pre-trained on ImageNet \cite{russakovsky2015imagenet} which also has been augmented with FiLM layers \cite{perez2018film}. The parameters $\{{\gamma_j}, {\beta_j}\}_{j=1}^4$ of the FiLM layers can scale and shift the extracted features at each layer of the ResNet18, allowing the feature extractor to focus and disregard different features on a task-by-task basis. 
A feature adaptation module $\psi_\phi^f$ is trained to produce $\{{\boldsymbol{\gamma}_j}, \boldsymbol{\beta}_j\}_{j=1}^4$ based on the support examples $\mathcal{S}^{\tau}$ provided for the task.

The feature extractor adaptation module $\psi_\phi^f$ consists of two stages:~support set encoding followed by film layer parameter production. The set encoder $g_{\phi}(\cdot)$, parameterized by a deep neural network, produces a permutation invariant task representation $g_{\phi}(\mathcal{S}^\tau)$ based on the support images $\mathcal{S}^\tau$. This task representation is then passed to $\psi_\phi^j$ which then produces the FiLM parameters $\{{\boldsymbol{\gamma}_j}, {\boldsymbol{\beta}_j}\}$ for each block $j$ in the ResNet. Once the FiLM parameters have been set, the feature extractor has been adapted to the task.  We use $f_\theta^\tau$ to denote the feature extractor adapted to task $\tau$. The CNAPS paper \cite{requeima2019fast}  also proposes an auto-regressive adaptation method which conditions each adaptor $\psi_\phi^j$ on the output of the previous adapter $\psi_\phi^{j-1}$. We refer to this variant as AR-CNAPS but for conciseness we omit the details of this architecture here, and instead refer the interested reader to \cite{requeima2019fast} or to Appendix \ref{appendix:ar-cnaps} for a brief overview.

Classification in CNAPS is performed by a task-adapted linear classifier where the class probabilities for a query image $\vx_i^*$ are computed as $\text{softmax}(\mathbf{W} f_\theta^\tau(\vx_i^*) + \mathbf{b})$.
The classification weights $\mathbf{W}$ and biases $\mathbf{b}$ are produced by the classifier adaptation network $\psi_\phi^c$ forming $[\mathbf{W}, \mathbf{b}] = [\psi_\phi^c(\vmu_1)\;\psi_\phi^c(\vmu_2)\; \ldots\; \psi_\phi^c(\vmu_K)]^T$
where for each class $k$ in the task, the corresponding row of classification weights is produced by $\psi_\phi^c$ from the class mean $\vmu_k$. The class mean $\vmu_k$ is obtained by mean-pooling the feature vectors of the support examples for class $k$ extracted by the adapted feature extractor $f_\theta^\tau$. A visual overview of the CNAPS adapted classifier architecture is shown in Figure \ref{cnaps-vs-us-direct-comparison}, bottom left, red.

\begin{figure}%
    \centering
    \subfloat[Euclidean Norm]{{\includegraphics[width=4cm]{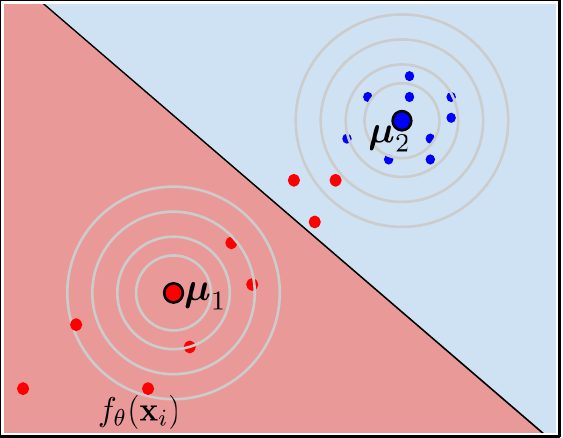} }}%
    \subfloat[Mahalanobis Distance]{{\includegraphics[width=4cm]{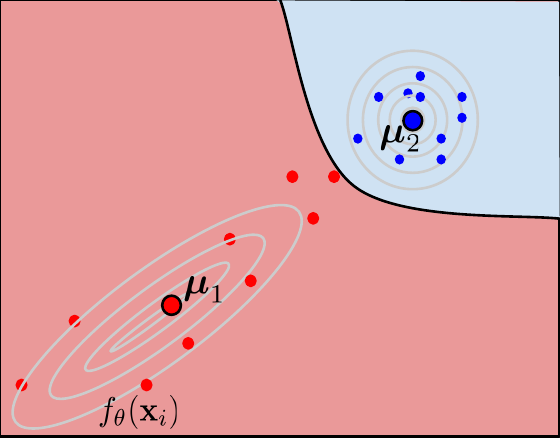} }}%
    \vspace{-0.1in}
    \caption{\textbf{Problematic nature of the unit-normal assumption:} The Euclidean Norm (left) assumes embedded image features $f_\theta(\vx_i)$ are distributed around class means $\vmu_k$ according to a unit normal. The Mahalanobis distance (right) considers cluster variance when forming decision boundaries, indicated by the background colour.}%
    \label{fig:gaussian}%
\end{figure}

\subsection{Simple CNAPS}
\label{sec:simple-cnaps}
In Simple CNAPS, we also use the same pre-trained ResNet18 for feature extraction with the same adaptation module $\psi_\phi^f$, although, because of the classifier architecture we use, it becomes trained to do something different than it does in CNAPS. This choice, like for CNAPS, allows for a task-specific adaptation of the feature extractor. Unlike CNAPS, we directly compute
\begin{equation}
    p(y_i^* = k|f_\theta^\tau(\vx_i^*), \mathcal{S}^\tau) = \text{softmax}(-d_k(f_\theta^\tau(\vx_i^*), \vmu_k)) \\
    \label{probability-calculation-simple-cnaps}
\end{equation}
using a deterministic, fixed $d_k$  
\begin{align}
    d_k(\vx, \vy) = \frac{1}{2}(\vx - \vy)^T(\mathbf{Q}_{k}^{\tau})^{-1}(\vx - \vy).\label{mh-distance}
\end{align}
Here $\mathbf{Q}_{k}^{\tau}$ is a covariance matrix specific to the task and class. 

As we cannot know the value of $\mathbf{Q}_{k}^{\tau}$ ahead of time, it must be estimated from the feature embeddings of the task-specific support set.   As the number of examples in any particular support set is likely to be much smaller than the dimension of the feature space, we use a regularized estimator
\begin{align}
   \mathbf{Q}_{k}^{\tau} &= \lambda_k^\tau \mathbf{\Sigma}_k^\tau+ (1 - \lambda_k^\tau )\mathbf{\Sigma}^{\tau} + \beta I.\label{cov-regularized-task-cov}
\end{align}
formed from a convex combination of the class-within-task and all-classes-in-task covariance matrices $\mathbf{\Sigma}_k^\tau$ and $\mathbf{\Sigma}^{\tau}$ respectively.

We estimate the class-within-task covariance matrix $\mathbf{\Sigma}_k^\tau$ using the feature embeddings $f^\tau_\theta(\vx_i)$ of all $\vx_i \in \mathcal{S}^{\tau}_k$ where $\mathcal{S}^{\tau}_k$ is the set of examples in $\mathcal{S}^{\tau}$ with class label $k$.
\begin{align*}
\mathbf{\Sigma}_k^\tau&=\frac{1}{|\mathcal{S}^\tau_k|-1}\sum_{(\vx_i,y_i)\in\mathcal{S}^\tau_k}(f_\theta^\tau(\vx_i)-\vmu_k)(f_\theta^\tau(\vx_i)-\vmu_k)^T.
\end{align*}
If the number of support instance of that class is one, i.e.~$|\mathcal{S}^\tau_k|$ = 1, then 
we define $\mathbf{\Sigma}_k^\tau$ to be the zero matrix of the appropriate size.
The all-classes-in-task covariance $\mathbf{\Sigma}^{\tau}$ is estimated in the same way as the class-within-task except that it uses \textit{all} the support set examples $\vx_i \in \mathcal{S}^\tau$ regardless of their class.

We choose a particular, deterministic scheme for computing the weighting of class and task specific covariance estimates, $\lambda_k^\tau = |\mathcal{S}_k^\tau| / (|\mathcal{S}_k^\tau| + 1)$. 
This choice means that in the case of a single labeled instance for class in the support set, a single ``shot,'' $\mathbf{Q}_{k}^{\tau} = 0.5\mathbf{\Sigma}_k^\tau+ 0.5\mathbf{\Sigma}^{\tau} + \beta \mathbf{I}$.  This can be viewed as increasing the strength of the regularization parameter $\beta$ relative to the task covariance $\mathbf{\Sigma}^{\tau}$. When $|\mathcal{S}_k^\tau| = 2$, $\lambda_k^\tau$ becomes $2/3$ and $\mathbf{Q}_{k}^{\tau}$ only partially favors the class-level covariance over the all-class-level covariance.  In a high-shot setting, $\lambda_k^\tau$ tends to $1$ and $\mathbf{Q}_{k}^{\tau}$ mainly consists of the class-level covariance. The intuition behind this formula for $\lambda_k^\tau$ is that the higher the number of shots, the better the class-within-task covariance estimate gets, and the more $\mathbf{Q}_{k}^{\tau}$ starts to look like $\mathbf{\Sigma}^\tau_k$. We considered other ratios and making $\lambda_k^\tau$'s learnable parameters, but found that out of all the considered alternatives the simple deterministic ratio above produced the best results. The architecture of the classifier in Simple CNAPS appears in Figure \ref{cnaps-vs-us-direct-comparison}, bottom-right, blue.

\section{Theory}
The class label probability calculation appearing in Equation~\ref{probability-calculation-simple-cnaps} corresponds to an equally-weighted exponential family mixture model as $\lambda \to 0$ \cite{snell2017prototypical},  where the exponential family distribution is uniquely determined by a regular Bregman divergence~\cite{banerjee2005clustering}
\begin{align}
    D_F(\vz, \vz') = F(\vz) - F(\vz') - \nabla F(\vz')^T(\vz - \vz')
\end{align}
for a differentiable and strictly convex function F. The squared Mahalanobis distance in Equation~\ref{mh-distance} is a Bregman divergence generated by the convex function $F(\vx) = \frac{1}{2} \vx^T\Sigma^{-1}\vx$ and corresponds to the multivariate normal exponential family distribution. When all $\mathbf{Q}_{k}^{\tau} \approx \Sigma^{\tau} + \beta I$, we can view the class probabilities in Equation~\ref{probability-calculation-simple-cnaps} as the ``responsibilities'' in a Gaussian mixture model
\begin{align}
    p(y_i^* = k|f_\theta^\tau(\vx_i^*), \mathcal{S}^\tau) = \frac{\pi_k \N(\vmu_k,\mathbf{Q}_{k}^{\tau})}{\sum_{k'}\pi_k' \N(\vmu_{k'}, \mathbf{Q}_{k}^{\tau})}\label{mh-normal-connection}
\end{align}
with equally weighted mixing coefficient $\pi_k = 1/k$. 

This perspective immediately highlights a problem with the squared Euclidean norm, used by a number of approaches as shown in Fig.~\ref{related-work-overview}. The Euclidean norm, which corresponds to the squared Mahalanobis distance with $\mathbf{Q}_k^\tau = \mathbf{I}$, implicitly assumes each cluster is distributed according to a unit normal, as seen in Figure \ref{fig:gaussian}. By contrast, the squared Mahalanobis distance considers cluster covariance when computing distances to the cluster centers.

\section{Experiments} \label{experiments-section}
We evaluate Simple CNAPS on the Meta-Dataset \cite{triantafillou2019meta} family of datasets, demonstrating improvements compared to nine baseline methodologies including the current SoTA, CNAPS. Benchmark results reported come from \cite{triantafillou2019meta, requeima2019fast}.

\subsection{Datasets}

\noindent
\textbf{Meta-Dataset} \cite{triantafillou2019meta} is a benchmark for few-shot learning encompassing $10$ labeled image datasets: ILSVRC-2012 (ImageNet) \cite{russakovsky2015imagenet}, Omniglot \cite{lake2015human}, FGVC-Aircraft (Aircraft) \cite{maji2013fine}, CUB-200-2011 (Birds) \cite{wah2011caltech}, Describable Textures (DTD) \cite{cimpoi2014describing}, QuickDraw \cite{jongejan2016quick}, FGVCx Fungi (Fungi)  \cite{fungi2018schroeder}, VGG Flower (Flower) \cite{nilsback2008automated}, Traffic Signs (Signs) \cite{houben2013detection} and MSCOCO \cite{lin2014microsoft}. In keeping with prior work, we report results using the first 8 datasets for training, reserving Traffic Signs and MSCOCO for ``out-of-domain'' performance evaluation. Additionally, from the eight training datasets used for training, some classes are held out for testing, to evaluate ``in-domain'' performance. Following \cite{requeima2019fast}, we extend the out-of-domain evaluation with 3 more datasets: MNIST \cite{lecun-mnisthandwrittendigit-2010}, CIFAR10 \cite{Krizhevsky09learningmultiple} and CIFAR100 \cite{Krizhevsky09learningmultiple}. We report results using standard test/train splits and benchmark baselines provided by \cite{triantafillou2019meta}, but, importantly, we have cross-validated our critical empirical claims using different test/train splits and our results are robust across folds (see Appendix \ref{appendix:cross-validation}). For details on task generation, distribution of shots/ways and hyperparameter settings, see Appendix \ref{appendix:experimental-details}.

\vspace{0.07in}
\noindent
\textbf{Mini/tieredImageNet} \cite{DBLP:journals/corr/abs-1803-00676-tieredimagenet,vinyals2016matching} are two smaller but more widely used benchmarks that consist of subsets of ILSVRC-2012 (ImageNet) \cite{russakovsky2015imagenet} with 100 classes (60,000 images) and 608 classes (779,165 images) respectively. For comparison to more recent work \cite{DBLP:journals/corr/abs-1804-09458-dynamic,DBLP:journals/corr/abs-1805-10002-tpn, NIPS2018_7352-tadam, DBLP:journals/corr/abs-1807-05960-leo} for which Meta-Dataset evaluations are not available, we use mini/tieredImageNet. Note that in the mini/tieredImageNet setting, all tasks are of the same pre-set number of classes and number of support examples per class, making learning comparatively easier.

\subsection{Results}
\label{sec:experiments-results}

\noindent
\textbf{Reporting format:} Bold indicates best performance on each dataset while underlines indicate statistically significant improvement over baselines. Error bars represent a 95\% confidence interval over tasks.

\begin{table*}[t]
    \centering
    \small
    \begin{tabular}{l|cccccccccc}
        & \multicolumn{8}{c}{In-Domain Accuracy (\%)}\\
        Model & ImageNet & Omniglot & Aircraft & Birds & DTD & QuickDraw & Fungi & Flower \\
        \hline
        MAML \cite{finn2017model} & 32.4\textpm1.0 & 71.9\textpm1.2 & 52.8\textpm0.9 & 47.2\textpm1.1 & 56.7\textpm0.7 & 50.5\textpm1.2 & 21.0\textpm1.0 & 70.9\textpm1.0 \\
        RelationNet \cite{sung2018learning} & 30.9\textpm0.9 & 86.6\textpm0.8 & 69.7\textpm0.8 & 54.1\textpm1.0 & 56.6\textpm0.7 & 61.8\textpm1.0 & 32.6\textpm1.1 & 76.1\textpm0.8 \\
        k-NN \cite{triantafillou2019meta} & 38.6\textpm0.9 & 74.6\textpm1.1 & 65.0\textpm0.8 & 66.4\textpm0.9 & 63.6\textpm0.8 & 44.9\textpm1.1 & 37.1\textpm1.1 & 83.5\textpm0.6 \\
        MatchingNet \cite{vinyals2016matching} & 36.1\textpm1.0 & 78.3\textpm1.0 & 69.2\textpm1.0 & 56.4\textpm1.0 & 61.8\textpm0.7 & 60.8\textpm1.0 & 33.7\textpm1.0 & 81.9\textpm0.7 \\
        Finetune \cite{DBLP:journals/corr/YosinskiCBL14-finetune} & 43.1\textpm1.1 & 71.1\textpm1.4 & 72.0\textpm1.1 & 59.8\textpm1.2 & 69.1\textpm0.9 & 47.1\textpm1.2 & 38.2\textpm1.0 & 85.3\textpm0.7 \\
        ProtoNet \cite{snell2017prototypical} & 44.5\textpm1.1 & 79.6\textpm1.1 & 71.1\textpm0.9 & 67.0\textpm1.0 & 65.2\textpm0.8 & 64.9\textpm0.9 & 40.3\textpm1.1 & 86.9\textpm0.7 \\
        ProtoMAML \cite{triantafillou2019meta} & 47.9\textpm1.1 & 82.9\textpm0.9 & 74.2\textpm0.8 & 70.0\textpm1.0 & 67.9\textpm0.8 & 66.6\textpm0.9 & 42.0\textpm1.1 & 88.5\textpm0.7 \\
        CNAPS \cite{requeima2019fast} & 51.3\textpm1.0 & 88.0\textpm0.7 & 76.8\textpm0.8 & 71.4\textpm0.9 & 62.5\textpm0.7 & 71.9\textpm0.8 & \textbf{46.0\textpm1.1} & 89.2\textpm0.5 \\
        AR-CNAPS \cite{requeima2019fast} & 52.3\textpm1.0 & 88.4\textpm0.7 & 80.5\textpm0.6 & 72.2\textpm0.9 & 58.3\textpm0.7 & 72.5\textpm0.8 & \textbf{47.4\textpm1.0} & 86.0\textpm0.5 \\
        \hline
        Simple AR-CNAPS & \underline{\textbf{56.5\textpm1.1}} & \underline{\textbf{91.1\textpm0.6}} & \underline{81.8\textpm0.8} & \underline{\textbf{74.3\textpm0.9}} & \underline{\textbf{72.8\textpm0.7}} & \underline{\textbf{75.2\textpm0.8}} & \textbf{45.6\textpm1.0} & \underline{\textbf{90.3\textpm0.5}} \\
        Simple CNAPS & \underline{\textbf{58.6\textpm1.1}} & \underline{\textbf{91.7\textpm0.6}} & \underline{\textbf{82.4\textpm0.7}} & \underline{\textbf{74.9\textpm0.8}} & 67.8\textpm0.8 & \underline{\textbf{77.7\textpm0.7}} & \textbf{46.9\textpm1.0} & \underline{\textbf{90.7\textpm0.5}} \\
    \end{tabular}
    \vspace{-0.1in}
    \caption{In-domain few-shot classification accuracy of Simple CNAPS and Simple AR-CNAPS compared to the baselines. With the exception of (AR-)CNAPS where the reported results are from \cite{requeima2019fast}, all other benchmarks are reported from \cite{triantafillou2019meta}.}
    \label{in-domain-results-film-cnaps}
    \vspace{-0.1in}
\end{table*}{}

\begin{table*}[t]
    \centering
    \small
    \begin{tabular}{l|ccccc|ccc}
        {} & \multicolumn{5}{c}{Out-of-Domain Accuracy (\%)} & \multicolumn{3}{c}{Average Accuracy (\%)} \\
        Model & Signs & MSCOCO & MNIST & CIFAR10 & CIFAR100 & In-Domain & Out-Domain & Overall \\
        \hline
        MAML \cite{finn2017model} & 34.2\textpm1.3 & 24.1\textpm1.1 & NA & NA & NA & 50.4\textpm1.0 & 29.2\textpm1.2 & 46.2\textpm1.1 \\
        RelationNet \cite{sung2018learning} & 37.5\textpm0.9 & 27.4\textpm0.9 & NA & NA & NA & 58.6\textpm0.9 & 32.5\textpm0.9 & 53.3\textpm0.9 \\
        k-NN \cite{triantafillou2019meta} & 40.1\textpm1.1 & 29.6\textpm1.0 & NA & NA & NA & 59.2\textpm0.9 & 34.9\textpm1.1 & 54.3\textpm0.9 \\
        MatchingNet \cite{vinyals2016matching} & 55.6\textpm1.1 & 28.8\textpm1.0 & NA & NA & NA & 59.8\textpm0.9 & 42.2\textpm1.1 & 56.3\textpm1.0 \\
        Finetune \cite{DBLP:journals/corr/YosinskiCBL14-finetune} & 66.7\textpm1.2 & 35.2\textpm1.1 & NA & NA & NA & 60.7\textpm1.1 & 51.0\textpm1.2 & 58.8\textpm1.1 \\
        ProtoNet \cite{snell2017prototypical} & 46.5\textpm1.0 & 39.9\textpm1.1 & 74.3\textpm0.8 & 66.4\textpm0.7 & 54.7\textpm1.1 & 64.9\textpm1.0 & 56.4\textpm0.9 & 61.6\textpm0.9 \\
        ProtoMAML \cite{triantafillou2019meta} & 52.3\textpm1.1 & 41.3\textpm1.0 & NA & NA & NA & 67.5\textpm0.9 & 46.8\textpm1.1 & 63.4\textpm0.9 \\
        CNAPS \cite{requeima2019fast} & 60.1\textpm0.9 & 42.3\textpm1.0 & 88.6\textpm0.5 & 60.0\textpm0.8 & 48.1\textpm1.0 & 69.6\textpm0.8 & 59.8\textpm0.8 & 65.9\textpm0.8 \\
        AR-CNAPS \cite{requeima2019fast} & 60.2\textpm0.9 & 42.9\textpm1.1 & 92.7\textpm0.4 & 61.5\textpm0.7 & 50.1\textpm1.0 & 69.7\textpm0.8 & 61.5\textpm0.8 & 66.5\textpm0.8 \\
        \hline
        Simple AR-CNAPS & \underline{\textbf{74.7\textpm0.7}} & 44.3\textpm1.1 & \underline{\textbf{95.7\textpm0.3}} & \underline{69.9\textpm0.8} & 53.6\textpm1.0 & \underline{\textbf{73.5\textpm0.8}} & \underline{67.6\textpm0.8} & \underline{\textbf{71.2\textpm0.8}} \\
        Simple CNAPS& \underline{\textbf{73.5\textpm0.7}} & \underline{\textbf{46.2\textpm1.1}} & \underline{93.9\textpm0.4} & \underline{\textbf{74.3\textpm0.7}} & \underline{\textbf{60.5\textpm1.0}} & \underline{\textbf{73.8\textpm0.8}} & \underline{\textbf{69.7\textpm0.8}} & \underline{\textbf{72.2\textpm0.8}} \\
    \end{tabular}
    \vspace{-0.15in}
    \caption{Middle) Out-of-domain few-shot classification accuracy of Simple CNAPS and Simple AR-CNAPS compared to the baselines. Right) In-domain, out-of-domain and overall mean classification accuracy of Simple CNAPS and Simple AR-CNAPS compared to the baselines. With the exception of CNAPS and AR-CNAPS where the reported results come from \cite{requeima2019fast}, all other benchmarks are reported directly from \cite{triantafillou2019meta}.}
    \label{out-domain-results-and-averages-film-cnaps}
    \vspace{-0.1in}
\end{table*}{}
\vspace{0.07in}
\noindent
\textbf{In-domain performance:} The in-domain results for Simple CNAPS and Simple AR-CNAPS, which uses the autoregressive feature extraction adaptor, are shown in Table \ref{in-domain-results-film-cnaps}. Simple AR-CNAPS outperforms previous SoTA on 7 out of the 8 datasets while matching past SoTA on FGVCx Fungi (Fungi). Simple CNAPS outperforms baselines on 6 out of 8 datasets while matching performance on FGVCx Fungi (Fungi) and Describable Textures (DTD). Overall, in-domain performance gains are considerable in the few-shot domain with 2-6\% margins. Simple CNAPS achieves an average 73.8\% accuracy on in-domain few-shot classification, a 4.2\% gain over CNAPS, while Simple AR-CNAPS achieves 73.5\% accuracy, a 3.8\% gain over AR-CNAPS.

\vspace{0.07in}
\noindent
\textbf{Out-of-domain performance:} As shown in Table \ref{out-domain-results-and-averages-film-cnaps}, Simple CNAPS and Simple AR-CNAPS produce substantial gains in performance on out-of-domain datasets, each exceeding the SoTA baseline. With an average out-of-domain accuracy of 69.7\% and 67.6\%, Simple CNAPS and Simple AR-CNAPS outperform SoTA by 8.2\% and 7.8\%.  This means that Simple CNAPS/AR-CNAPS generalizes to out-of-domain datasets better than baseline models. Also, Simple AR-CNAPS under-performs Simple CNAPS, suggesting that the auto-regressive feature adaptation approach may overfit to the domain of datasets it has been trained on.

\vspace{0.07in}
\noindent
\textbf{Overall performance:} Simple CNAPS achieves the best overall classification accuracy at 72.2\% with Simple AR-CNAPS trailing very closely at 71.2\%.  Since the overall performance of the two variants are statistically indistinguishable, we recommend Simple CNAPS over Simple AR-CNAPS as it has fewer parameters. 

\begin{figure}
    \centering
    \includegraphics[width=3.1in]{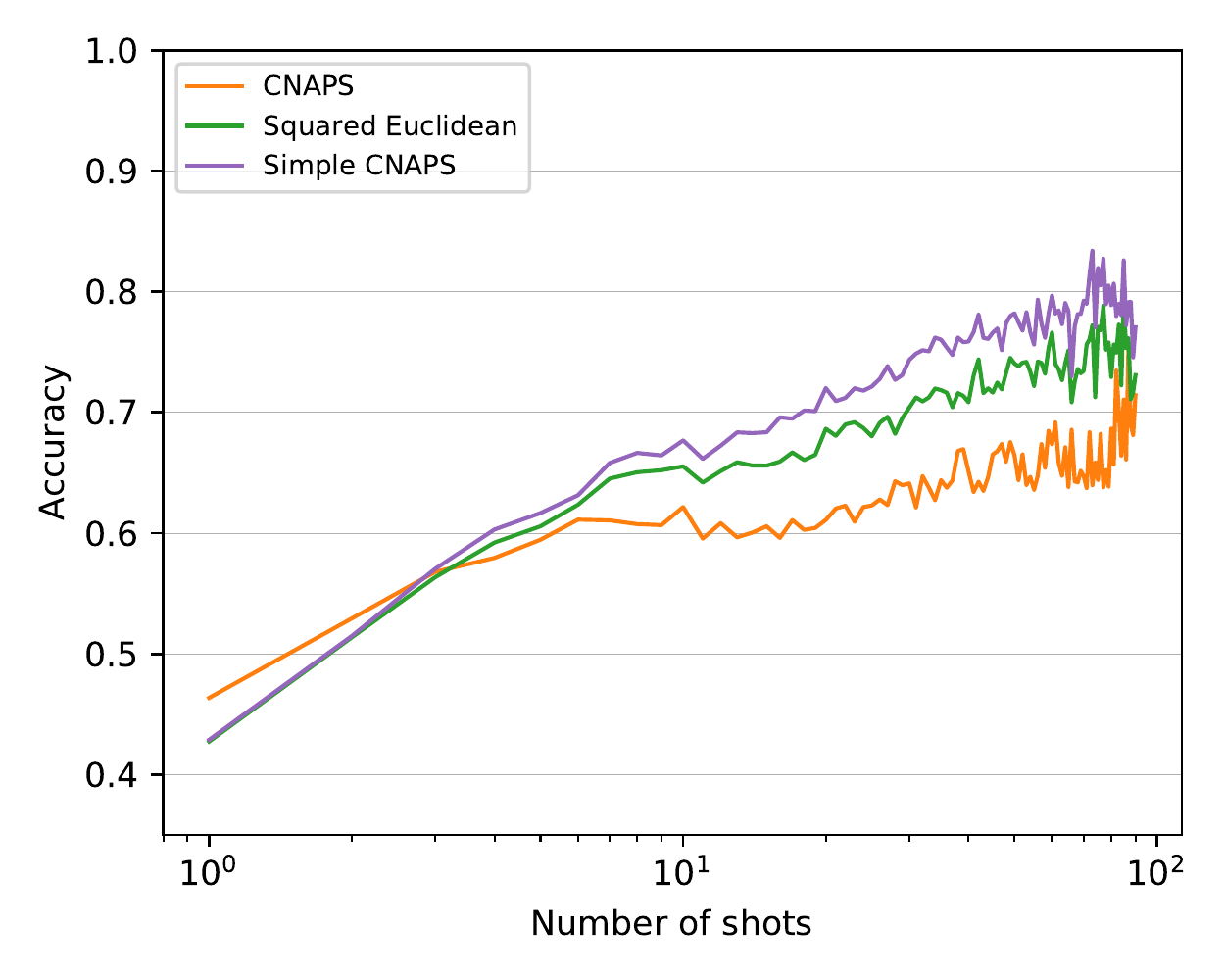}
    \vspace{-0.2in}
    \caption{\textbf{Accuracy vs. Shots:} Average number of support examples (in log scale) per class v/s accuracy. TFor each class in each of the 7,800 sampled Meta-Dataset tasks (13 datasets, 600 tasks each) used at test time, the classification accuracy on the class' query examples was obtained. These class accuracies were then grouped according to the class shot, averaged and plotted to show how accuracy of CNAPS, $L_2^2$ and Simple-CNAPS scale with higher shots.}
    \vspace{-0.1in}
    \label{mah-shots-graph}
\end{figure}

\begin{figure}[ht]
    \centering
    \includegraphics[width=3.1in]{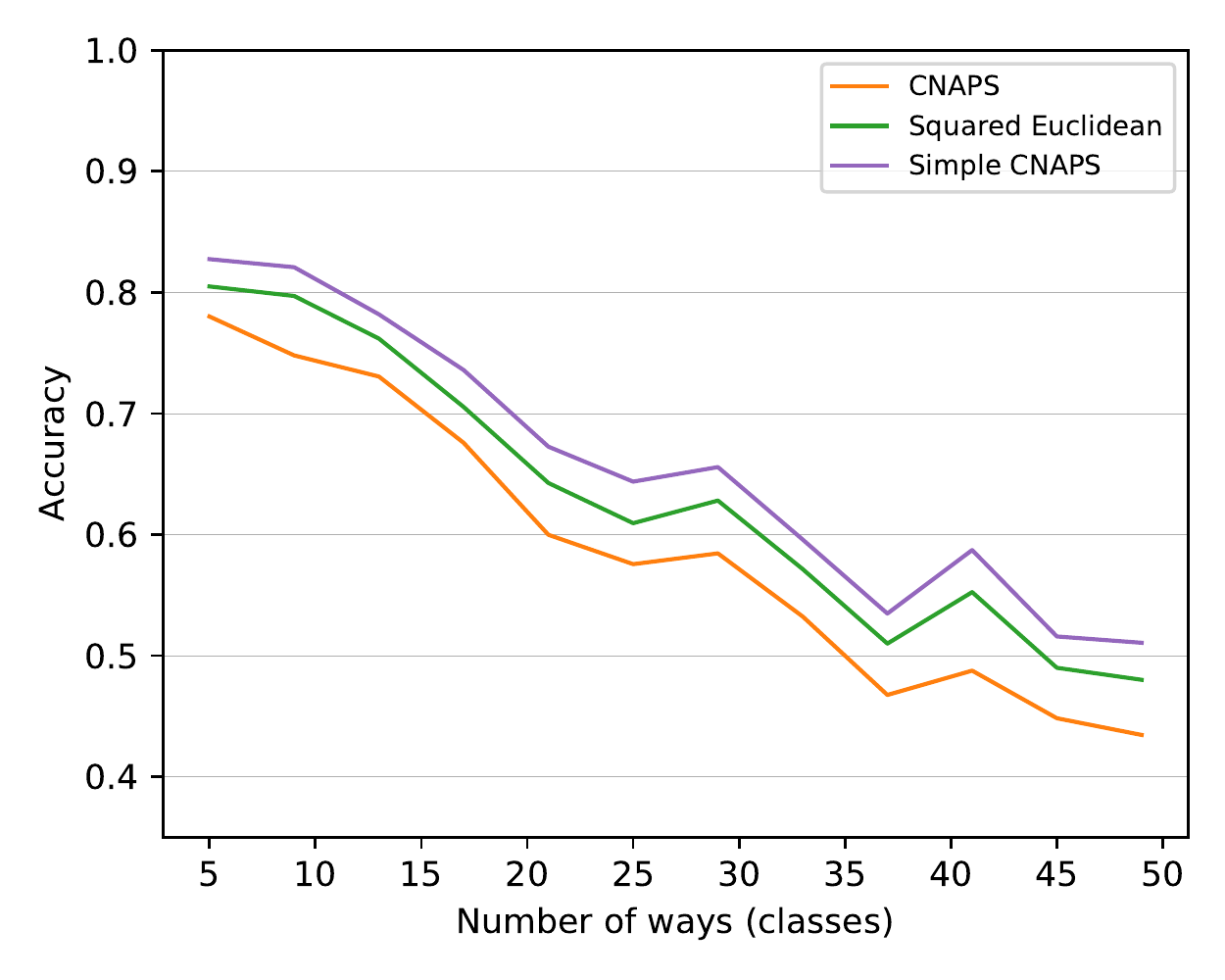}
     \vspace{-0.2in}
    \caption{\textbf{Accuracy vs. Ways:} Number of ways (classes in the task) v/s accuracy. Tasks in the test set are grouped together by number of classes. The accuracies are averaged to obtain a value for each count of class.}
    \vspace{-0.1in}
    \label{mah-ways-graphs}
\end{figure}

\begin{table*}[t]
    \centering
    \small
    \begin{tabular}{l|cccccccccc}
        & \multicolumn{8}{c}{In-Domain Accuracy (\%)}\\
        Metric & ImageNet & Omniglot & Aircraft & Birds & DTD & QuickDraw & Fungi & Flower \\
        \hline
        Negative Dot Product & 48.0\textpm1.1 & 83.5\textpm0.9 & 73.7\textpm0.8 & 69.0\textpm1.0 & 66.3\textpm0.6 & 66.5\textpm0.9 & 39.7\textpm1.1 & 88.6\textpm0.5 \\
        Cosine Similarity & 51.3\textpm1.1 & 89.4\textpm0.7 & 80.5\textpm0.8 & 70.9\textpm1.0 & \textbf{69.7\textpm0.7} & 72.6\textpm0.9 & 41.9\textpm1.0 & 89.3\textpm0.6 \\
        Absolute Distance ($L_1$) & 53.6\textpm1.1 & \textbf{90.6\textpm0.6} & 81.0\textpm0.7 & \textbf{73.2\textpm0.9} & 61.1\textpm0.7 & 74.1\textpm0.8 & \textbf{47.0\textpm1.0} & 87.3\textpm0.6 \\
        Squared Euclidean (${L_2}^2$) & 53.9\textpm1.1 & \textbf{90.9\textpm0.6} & \textbf{81.8\textpm0.7} & \textbf{73.1\textpm0.9} & 64.4\textpm0.7 & 74.9\textpm0.8 & \textbf{45.8\textpm1.0} & 88.8\textpm0.5 \\
        \hline
        Simple CNAPS -TR & \underline{\textbf{56.7\textpm1.1}} & \underline{\textbf{91.1\textpm0.7}} & \textbf{83.0\textpm0.7} & \textbf{74.6\textpm0.9} & \underline{\textbf{70.2\textpm0.8}} & \textbf{76.3\textpm0.9} & \textbf{46.4\textpm1.0} & \textbf{90.0\textpm0.6} \\
        Simple CNAPS & \underline{\textbf{58.6\textpm1.1}} & \underline{\textbf{91.7\textpm0.6}} & \textbf{82.4\textpm0.7} & \textbf{74.9\textpm0.8} & \underline{67.8\textpm0.8} & \underline{\textbf{77.7\textpm0.7}} & \textbf{46.9\textpm1.0} & \underline{\textbf{90.7\textpm0.5}} \\
    \end{tabular}
    \vspace{-0.1in}
    \caption{In-domain few-shot classification accuracy of Simple CNAPS compared to ablated alternatives of the negative dot product, absolute difference ($L_1$), squared Euclidean (${L_2}^2$) and removing task regularization ($\lambda_k^\tau = 1$) denoted by "-TR".}
    \label{in-domain-results}
    \vspace{-0.1in}
\end{table*}{}

\begin{table*}[t]
    \centering
    \small
    \begin{tabular}{l|ccccc|ccc}
        {} & \multicolumn{5}{c}{Out-of-Domain Accuracy (\%)} & \multicolumn{3}{c}{Average Accuracy (\%)} \\
        Metric & Signs & MSCOCO & MNIST & CIFAR10 & CIFAR100 & In-Domain & Out-Domain & Overall \\
        \hline
        Negative Dot Product & 53.9\textpm0.9 & 32.5\textpm1.0 & 86.4\textpm0.6 & 57.9\textpm0.8 & 38.8\textpm0.9 & 66.9\textpm0.9 & 53.9\textpm0.8 & 61.9\textpm0.9 \\
        Cosine Similarity & 65.4\textpm0.8 & 41.0\textpm1.0 & 92.8\textpm0.4 & 69.5\textpm0.8 & 53.6\textpm1.0 & 70.7\textpm0.9 & 64.5\textpm0.8 & 68.3\textpm0.8 \\
        Absolute Distance ($L_1$) & 66.4\textpm0.8 & 44.7\textpm1.0 & 88.0\textpm0.5 & 70.0\textpm0.8 & 57.9\textpm1.0 & 71.0\textpm0.8 & 65.4\textpm0.8 & 68.8\textpm0.8 \\
        Squared Euclidean (${L_2}^2$) & 68.5\textpm0.7 & 43.4\textpm1.0 & 91.6\textpm0.5 & 70.5\textpm0.7 & 57.3\textpm1.0 & 71.7\textpm0.8 & 66.3\textpm0.8 & 69.6\textpm0.8 \\
        \hline
        Simple CNAPS -TR & \underline{\textbf{74.1\textpm0.6}} & \underline{\textbf{46.9\textpm1.1}} & \underline{\textbf{94.8\textpm0.4}} & \underline{\textbf{73.0\textpm0.8}} & \textbf{59.2\textpm1.0} & \underline{\textbf{73.5\textpm0.8}} & \underline{\textbf{69.6\textpm0.8}} & \underline{\textbf{72.0\textpm0.8}} \\
        Simple CNAPS& \underline{\textbf{73.5\textpm0.7}} & \underline{\textbf{46.2\textpm1.1}} & \underline{93.9\textpm0.4} & \underline{\textbf{74.3\textpm0.7}} & \underline{\textbf{60.5\textpm1.0}} & \underline{\textbf{73.8\textpm0.8}} & \underline{\textbf{69.7\textpm0.8}} & \underline{\textbf{72.2\textpm0.8}} \\
    \end{tabular}
    \vspace{-0.1in}
    \caption{Middle) Out-of-domain few-shot classification accuracy of Simple CNAPS compared to ablated alternatives of the negative dot product, absolute difference ($L_1$), squared Euclidean (${L_2}^2$) and removing task regularization ($\lambda_k^\tau = 1$) denoted by "-TR". Right) In-domain, out-of-domain and overall mean classification accuracies of the ablated models.}
    \label{out-domain-results-and-averages}
\end{table*}{}

\begin{table}[t]
    \centering
    \small
    \begin{tabular}{l|cccc}
        {} & \multicolumn{2}{c}{\textit{mini}ImageNet} & \multicolumn{2}{c}{\textit{tiered}ImageNet} \\
        Model & 1-shot & 5-shot & 1-shot & 5-shot \\
        \hline
        ProtoNet \cite{snell2017prototypical} & 46.14 & 65.77 & 48.58 & 69.57 \\
        Gidariss et al. \cite{DBLP:journals/corr/abs-1804-09458-dynamic} & 56.20 & 73.00 & N/A & N/A \\
        TADAM \cite{NIPS2018_7352-tadam} & 58.50 & 76.70 & N/A & N/A \\
        TPN \cite{DBLP:journals/corr/abs-1805-10002-tpn} & 55.51 & 69.86 & 59.91 & 73.30 \\
        LEO \cite{DBLP:journals/corr/abs-1807-05960-leo} & 61.76 & 77.59 & 66.33 & 81.44 \\
        \hdashline
        CNAPS \cite{requeima2019fast} & 77.99 & 87.31 & 75.12 & 86.57 \\
        \hline
        Simple CNAPS  & \underline{\textbf{82.16}} & \underline{\textbf{89.80}} & \underline{\textbf{78.29}} & \underline{\textbf{89.01}} \\
    \end{tabular}
    \vspace{-0.1in}
    \caption{Accuracy (\%) compared to mini/tieredImageNet baselines. Performance measures reported for CNAPS and Simple CNAPS are averaged across 5 different runs.}
    \vspace{-0.15in}
    \label{mini-and-tiered-imagenet-results}
\end{table}{}

\vspace{0.07in}
\noindent
\textbf{Comparison to other distance metrics:}
\label{sec:experiments-ablation}
To test the significance of our choice of Mahalanobis distance, we substitute it within our architecture with other distance metrics - absolute difference ($L_1$), squared Euclidean ($L_2^2$), cosine similarity and negative dot-product.  Performance comparisons are shown in Table \ref{in-domain-results} and \ref{out-domain-results-and-averages}. We observe that using the Mahalanobis distance results in the best in-domain, out-of-domain, and overall average performance on all datasets.

\vspace{0.07in}
\noindent
\textbf{Impact of the task regularizer $\Sigma^\tau$:}
We also consider a variant of Simple CNAPS where all-classes-within-task covariance matrix $\Sigma^\tau$ is not included in the covariance regularization (denoted with the "-TR" tag). This is equivalent to setting $\lambda_k^\tau$ to 1 in Equation \ref{cov-regularized-task-cov}. As shown in Table \ref{out-domain-results-and-averages}, we observe that, while removing the task level regularizer only marginally reduces overall performance, the difference on individual datasets such as ImageNet can be large.

\vspace{0.07in}
\noindent
\textbf{Sensitivity to the number of support examples per class:}
Figure \ref{mah-shots-graph} shows how the overall classification accuracy varies as a function of the average number of support examples per class (shots) over all tasks. We compare Simple CNAPS, original CNAPS, and the $L_2^2$ variant of our method.  As expected, the average number of support examples per class is highly correlated with the performance.  All methods perform better with more labeled examples per support class, with Simple CNAPS performing substantially better as the number of shots increases.  The surprising discovery is that Simple CNAPS is effective even when the number of labeled instances is as low as four, suggesting both that even poor estimates of the task and class specific covariance matrices are helpful and that the regularization scheme we have introduced works remarkably well.

\vspace{0.07in}
\noindent
\textbf{Sensitivity to the number of classes in the task:} In Figure \ref{mah-ways-graphs}, we examine average accuracy as a function of the number of classes in the task. We find that,  irrespective of the number of classes in the task, we maintain accuracy improvement over both CNAPS and our $L_2^2$ variant.

\vspace{0.07in}
\noindent
\textbf{Accuracy on mini/tieredImageNet:} Table \ref{mini-and-tiered-imagenet-results} shows that Simple CNAPS outperforms recent baselines on all of the standard 1- and 5-shot 5-way classification tasks. These results should be interpreted with care as both CNAPS and Simple CNAPS use a ResNet18  \cite{DBLP:journals/corr/HeZRS15-resnet} feature extractor \emph{pre-trained on ImageNet}.  Like other models in this table, here Simple CNAPS was trained for these particular shot/way configurations.  That Simple CNAPS performs well here in the 1-shot setting,  improving even on CNAPS, suggests that Simple CNAPS is able to specialize to particular few-shot classification settings in addition to performing well when the number of shots and ways is unconstrained as it was in the earlier experiments.

\section{Discussion}
\label{sec:discussion}

Few shot learning is a fundamental task in modern AI research.  In this paper we have introduced a new  method for amortized few shot image classification which establishes a new SoTA performance benchmark by making a simplification to the current SoTA architecture.  Our specific architectural choice, that of deterministically estimating and using Mahalanobis distances for classification of task-adjusted class-specific feature vectors, seems to produce, via training, embeddings that generally allow for useful covariance estimates, even when the number of labeled instances, per task and class, is small.  The effectiveness of the Mahalanobis distance in feature space for distinguishing classes suggests connections to hierarchical regularization schemes \cite{pmlr-v27-salakhutdinov12a} that could enable performance improvements even in the zero-shot setting.  In the future, exploration of other Bregman divergences can be an avenue of potentially fruitful research. Additional enhancements in the form of data and task augmentation can also boost the performance.

\section{Acknowledgements}
We acknowledge the support of the Natural Sciences and Engineering Research Council of Canada (NSERC), the Canada Research Chairs (CRC) Program, the Canada CIFAR AI Chairs Program, Compute Canada, Intel, and DARPA under its D3M and LWLL programs.

\newpage

{\small
\bibliographystyle{ieee}
\bibliography{egpaper_final}
}

\newpage
\newpage
\clearpage

\begin{appendices}

\section{Experimental Setting}
\label{appendix:experimental-details}

\begin{figure}[!b]
    \centering
    \includegraphics[width=2.55in]{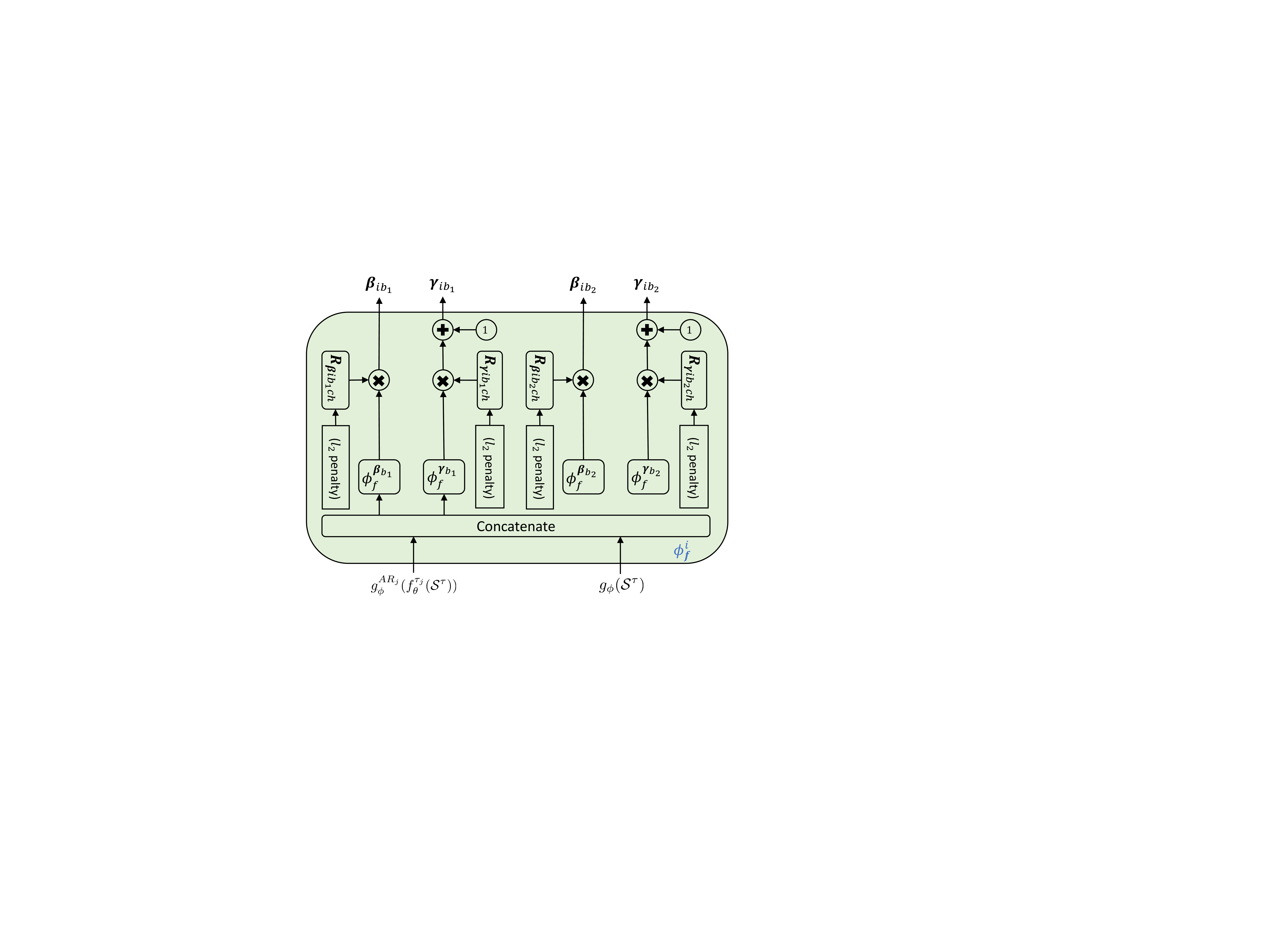}
    \vspace{-0.1in}
    \caption{\textbf{Architectural overview of the feature extractor adaptation network $\psi_\phi^f$:} Figure has been adapted from \cite{requeima2019fast} and showcases the neural architecture used for each adaptation module $\psi_\phi^j$ (corresponding to residual block $j$) in the feature extractor adaptation network $\psi_\phi^f$.}
    \label{adaptation-network-architecture}
    \vspace{-0.1in}
\end{figure}

\begin{figure}[t]
    \centering
    \small
    \subfloat[Number of Tasks vs. Ways]{{\includegraphics[width=4.0cm]{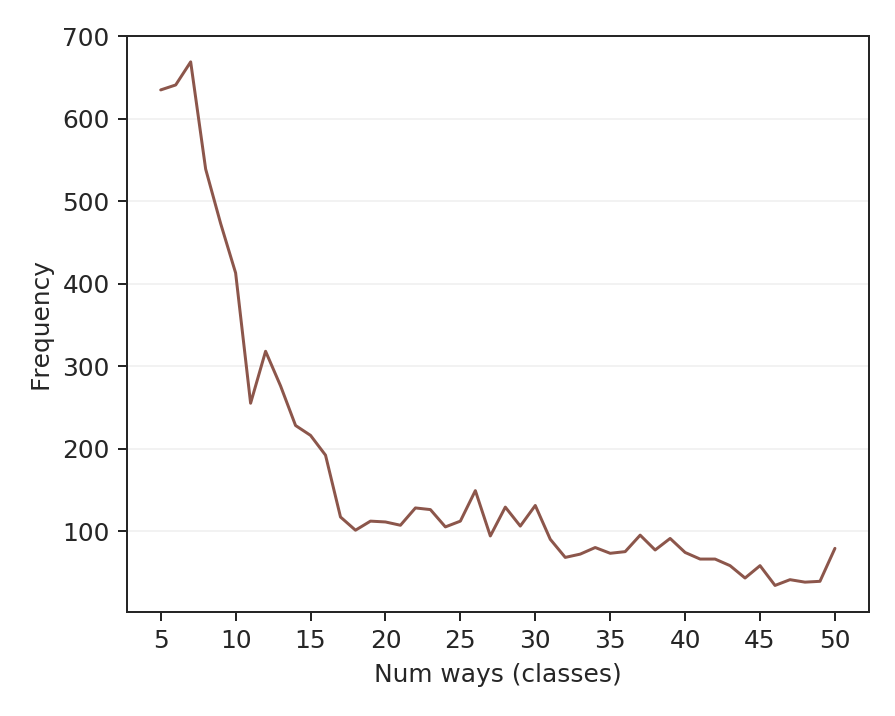} \label{fig:ways-vs-freq}}}%
    \subfloat[Number of Classes vs. Shots]{{\includegraphics[width=4.0cm]{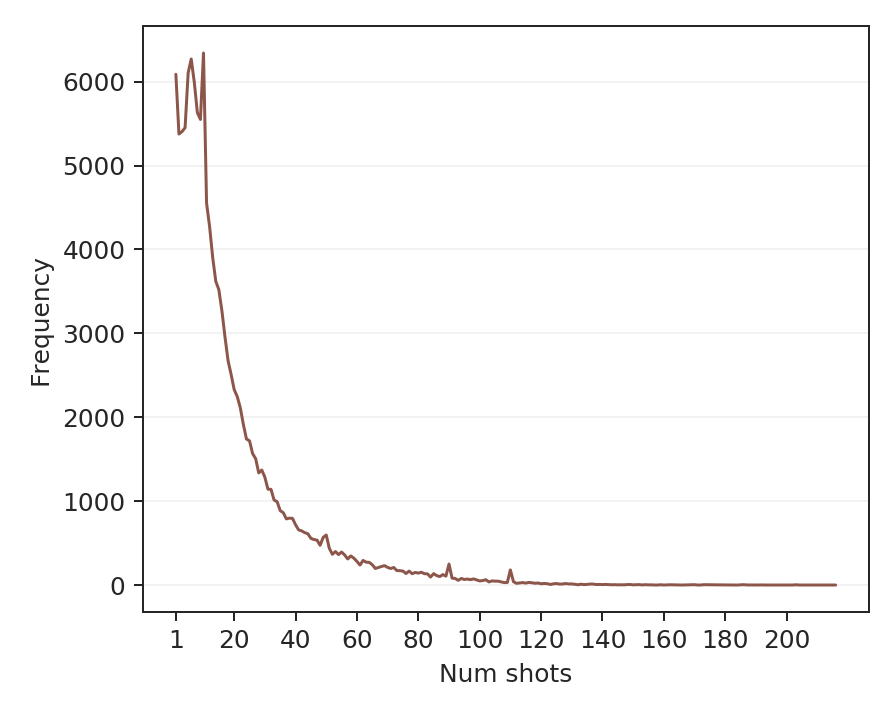} \label{fig:shots-vs-freq}}}%
    \vspace{-0.1in}
    \caption{\textbf{Test-time distribution of tasks:} a) Frequency of number of tasks as grouped by the number of classes in the tasks (ways). b) Frequency of the number of classes grouped by the number examples per class (shots). Both figures are for test tasks sampled when evaluating on the Meta-Dataset \cite{triantafillou2019meta}.}
    \vspace{-0.1in}
\end{figure}

Section 3.2 of \cite{triantafillou2019meta} explains the sampling procedure to generate tasks from the Meta-Dataset \cite{triantafillou2019meta}, used during both training and testing. This results in tasks with varying of number of shots/ways. Figure \ref{fig:ways-vs-freq} and \ref{fig:shots-vs-freq} show the ways/shots frequency graphs at test time. For evaluating on Meta-Dataset and mini/tiered-ImageNet datasets, we use \textit{episodic training} \cite{snell2017prototypical} to train models to remain consistent with the prior works \cite{finn2017model, requeima2019fast, snell2017prototypical, triantafillou2019meta}. We train for 110K tasks, 16 tasks per batch, totalling 6,875 gradient steps using Adam with learning rate of 0.0005. We validate (on 8 in-domain and 1 out-of-domain datasets) every 10K tasks, saving the best model/checkpoint for testing. Please visit the \href{https://github.com/peymanbateni/simple-cnaps}{Pytorch implementation of Simple CNAPS} for details.

\section{(Simple) CNAPS in Details}
\label{appendix:cnaps-details}

\subsection{Auto-Regressive CNAPS}
\label{appendix:ar-cnaps}

\begin{figure*}[t]
    \centering
    \includegraphics[width=6.7in]{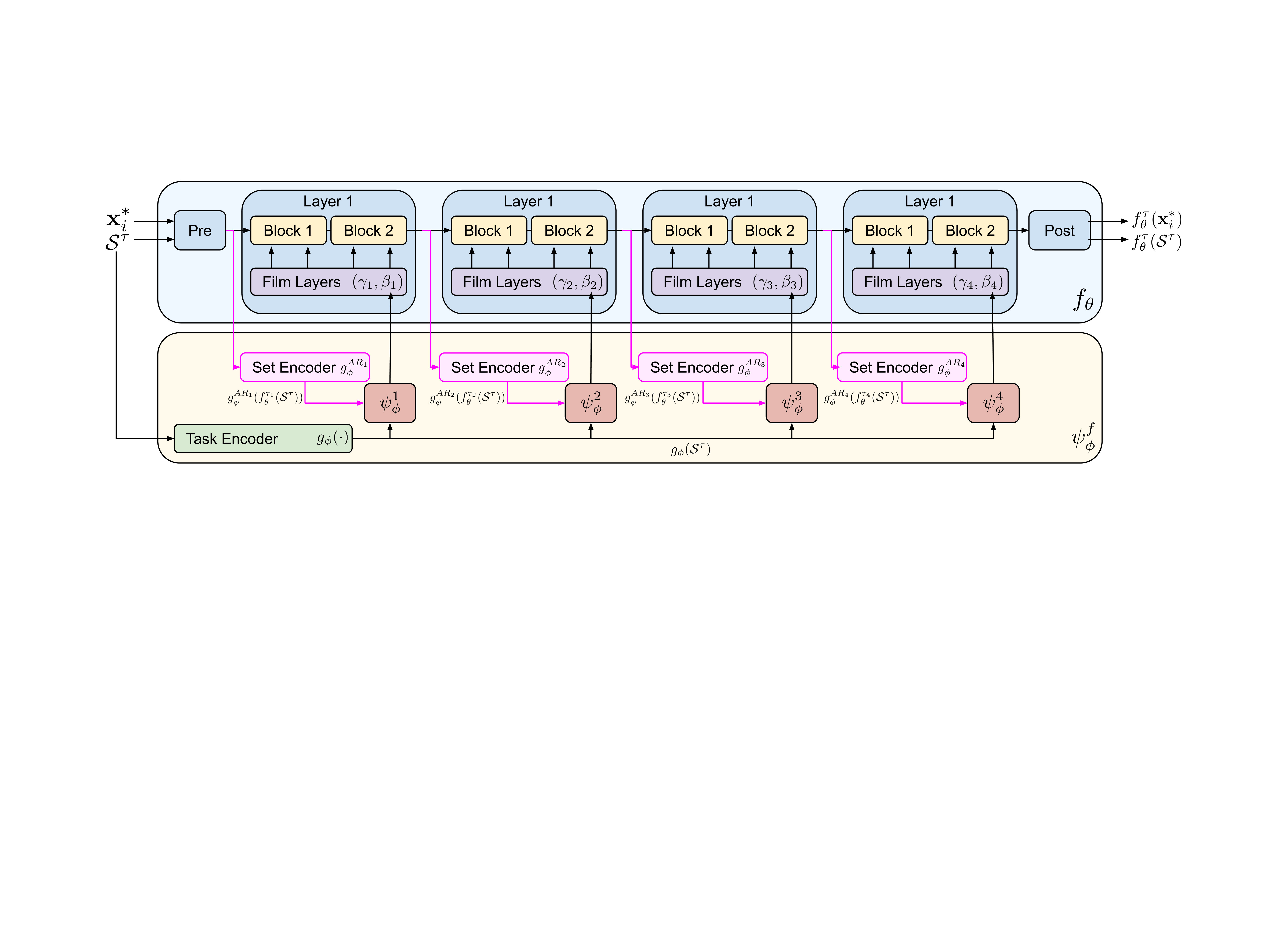}
    \vspace{-0.07in}
    \caption{\textbf{Overview of the auto-regresive feature extractor adaptation in CNAPS:} in addition to the structure shown in Figure \ref{cnaps-feature-extraction-overview}, AR-CNAPS takes advantage of a series of pre-block set encoders $g_\phi^{{AR}_j}$ to furthermore condition the output of each $\psi_\phi^j$ on the set representation $g_\phi^{{AR}_j}(f_\theta^{\tau_j}(\mathcal{S}^\tau))$. The set representation is formed by first adapting the previous blocks $1:j-1$, then pushing the support set $\mathcal{S}$ through the adapted blocks to form an auto-regressive adapted set representation at block $j$. This way, adaptive functions later in the pipeline are more explicitly aware of the changes made by the previous adaptation networks, and can adjust better accordingly.}
    \vspace{-0.1in}
    \label{auto-regressive-cnaps-feature-extraction-overview}
\end{figure*}

In \cite{requeima2019fast}, an additional auto-regressive variant for adapting the feature extractor is proposed, referred to as AR-CNAPS. As shown in Figure \ref{auto-regressive-cnaps-feature-extraction-overview}, AR-CNAPS extends CNAPS by introducing the block-level set encoder $g_\phi^{{AR}_j}$ at each block $j$. These set encoders use the output obtained by pushing the support $\mathcal{S}^\tau$ through all previous blocks $1:j-1$ to form the block level set representation $g_\phi^{{AR}_j}(f_\theta^{\tau_j}(\mathcal{S}^\tau))$. This representation is then subsequently used as input to the adaptation network $\psi_\phi^j$ in addition to the task representation $g_\phi(\mathcal{S}^\tau)$. This way the adaptation network is not just conditioned on the task, but is also aware of the potential changes in the previous blocks as a result of the adaptation being performed by the adaptation networks before it (\ie, $\psi_\phi^1:\psi_\phi^{j-1}$). The auto-regressive nature of AR-CNAPS allows for a more dynamic adaptation procedure that boosts performance in certain domains.

\subsection{FiLM Layers}
\label{appendix:cnaps-film}

\begin{figure*}[t]
    \centering
    \includegraphics{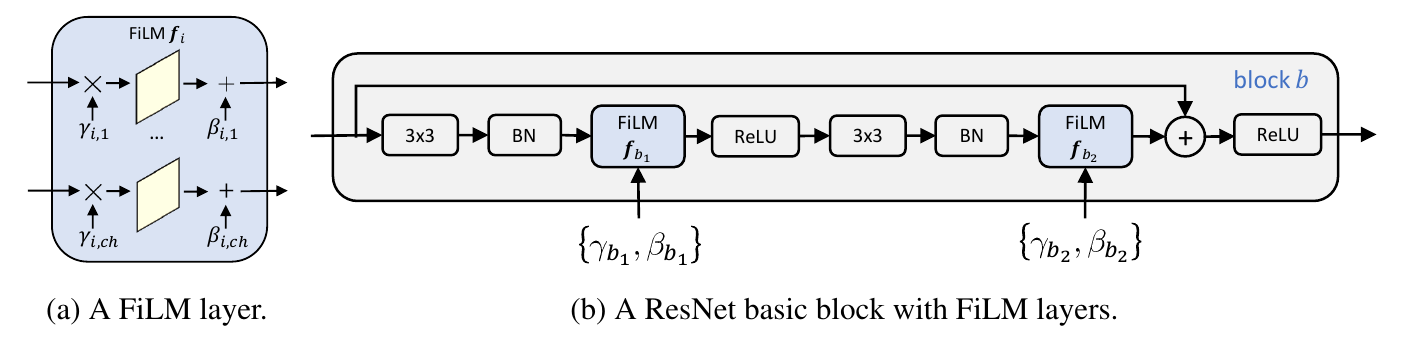}
    \vspace{-0.1in}
    \caption{\textbf{Overview of FiLM layers:} Figure is from
    \cite{requeima2019fast}. Left) FiLM layer operating a series of channels indexed by $ch$, scaling and shifting the feature channels as defined by the respective FiLM parameters $\gamma_{i,ch}$ and $\beta_{i, ch}$. Right) Placement of these FiLM modules within a ResNet18 \cite{DBLP:journals/corr/HeZRS15-resnet} basic block.}
    \vspace{-0.1in}
    \label{film-layers-overview}
\end{figure*}

Proposed by \cite{perez2018film}, Feature-wise Linear Modulation (FiLM) layers were used for visual question answering, where the feature extractor could be conditioned on the question. As shown in Figure \ref{film-layers-overview}, these layers are inserted within residual blocks, where the feature channels 
are scaled and linearly shifted using the respective FiLM parameters $\gamma_{i,ch}$ and $\beta_{i,ch}$. This can be extremely powerful in transforming the extracted feature space. In our work and \cite{requeima2019fast}, these FiLM parameters are conditioned on the support images in the task $\mathcal{S}^\tau$. This way, the adapted feature extractor $f_\theta^\tau$ is able to modify the feature space to extract the features that allow classes in the task to be distinguished most distinctly. This is in particular very powerful when the classification metric is changed to the Mahalanobis distance, as with a new objective, the feature extractor adaptation network $\psi_\phi^f$ is able to learn to extract better features (see difference between with and without $\psi_\phi^f$ in Table \ref{feature-extractor-adaptation-ablation} on CNAPS and Simple CNAPS).

\subsection{Network Architectures}
\label{appendix:cnaps-architectures}

\begin{figure}[t]
    \centering
    \includegraphics[width=2.5in]{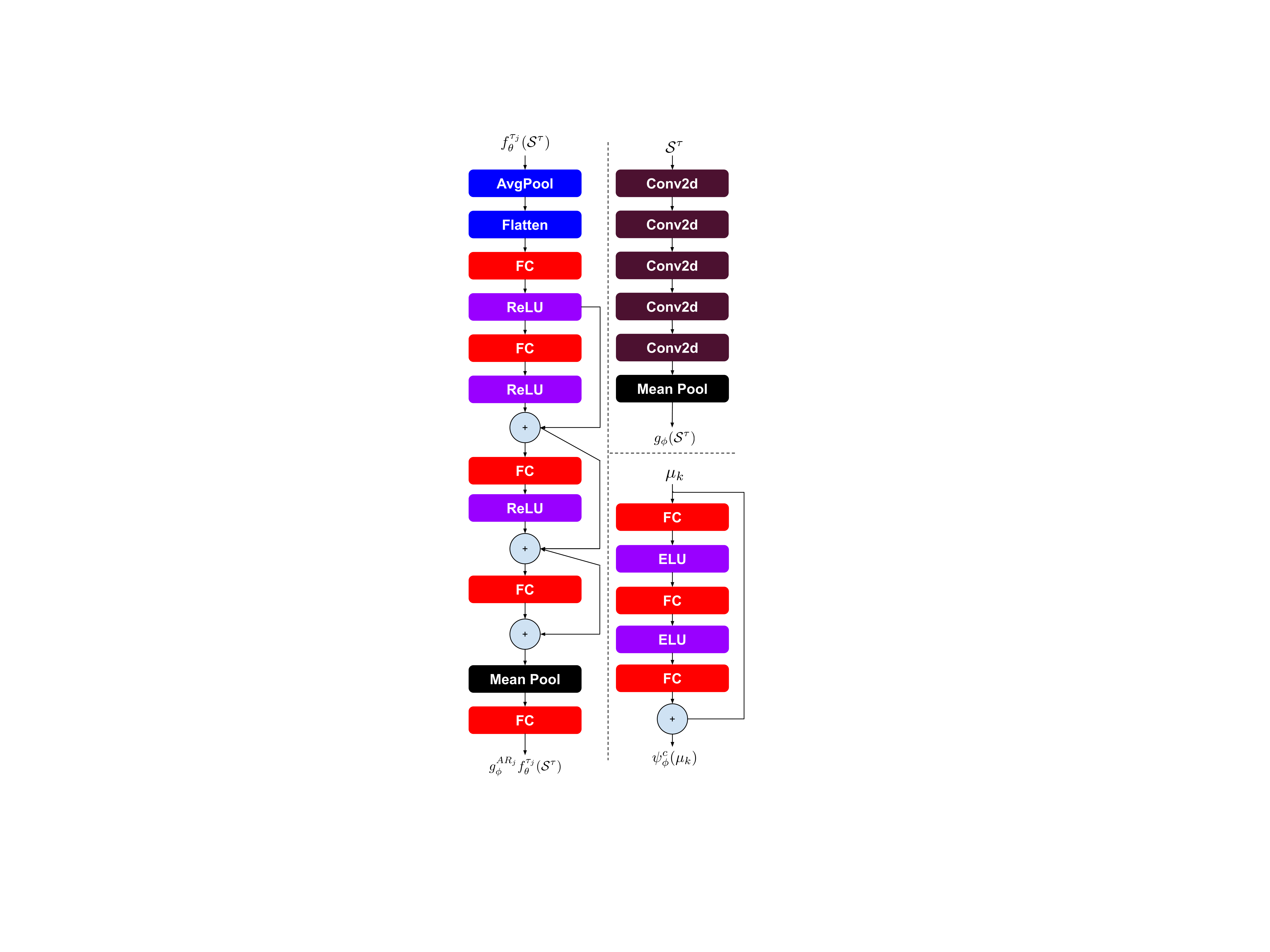}
    \vspace{-0.1in}
    \caption{\textbf{Overview of architectures used in (Simple) CNAPS:} a) Auto-regressive set encoder $g_\phi^{AR_j}$. Note that since this is conditioned on the channel outputs of the convolutional filter, it's not convolved any further. b) Task encoder $g_\phi$ that mean-pools convolutionally filtered support examples to produce the task representation. c) architectural overview of the classifier adaptation network $\psi_\phi^c$ consisting of a 3 layer MLP with a residual connection. Diagrams are based on Table E.8, E.9, and E.11 in \cite{requeima2019fast}.}
    \vspace{-0.1in}
    \label{set-encoders-and-classifier}
\end{figure}{}

\begin{table*}[t]
    \small
    \centering
    \begin{tabular}{c|cccccccccc}
        & \multicolumn{8}{c}{Classification Accuracy (\%)}\\
        Model & ILSVRC & Omniglot & Aircraft & CUB & DTD & QuickDraw & Fungi & Flower \\
        \hline
        CNAPS& 49.6\textpm1.1 & 87.2\textpm0.8 & 81.0\textpm0.7 & 69.7\textpm0.9 & 61.3\textpm0.7 & 72.0\textpm0.8 & *32.2\textpm1.0 & *70.9\textpm0.8   \\
        Simple CNAPS & \textbf{55.6\textpm1.1} & \textbf{90.9\textpm0.8} & 82.2\textpm0.7 & \textbf{75.4\textpm0.9} & \textbf{74.3\textpm0.7} & \textbf{75.5\textpm0.8} & \textbf{*39.9\textpm1.0} & \textbf{*88.0\textpm0.8}   \\
        \hline
        CNAPS & 50.3\textpm1.1 & 86.5\textpm0.8 & 77.1\textpm0.7 & 71.6\textpm0.9 & *64.3\textpm0.7 & *33.5\textpm0.9 & 46.4\textpm1.1 & 84.0\textpm0.6   \\
        Simple CNAPS & \textbf{58.1\textpm1.1} & \textbf{90.8\textpm0.8} & \textbf{83.8\textpm0.7} & \textbf{75.2\textpm0.9} & \textbf{*74.6\textpm0.7} & \textbf{*64.0\textpm0.9} & 47.7\textpm1.1 & \textbf{89.9\textpm0.6}    \\
        \hline
        CNAPS & 51.5\textpm1.1 & 87.8\textpm0.8 & *38.2\textpm0.8 & *58.7\textpm1.0 & 62.4\textpm0.7 & 72.5\textpm0.8 & 46.9\textpm1.1 & 89.4\textpm0.5   \\
        Simple CNAPS & \textbf{56.0\textpm1.1} & \textbf{91.1\textpm0.8} & \textbf{*66.6\textpm0.8} & \textbf{*68.0\textpm1.0} & \textbf{71.3\textpm0.7} & \textbf{76.1\textpm0.8} & 45.6\textpm1.1 & \textbf{90.7\textpm0.5}    \\
        \hline
        CNAPS & *42.4\textpm0.9 & *59.6\textpm1.4 & 77.2\textpm0.8 & 69.3\textpm0.9 & 62.9\textpm0.7 & 69.1\textpm0.8 & 40.9\textpm1.0 & 88.2\textpm0.5   \\
        Simple CNAPS & \textbf{*49.1\textpm0.9} & \textbf{*76.0\textpm1.4} & \textbf{83.0\textpm0.8} & \textbf{74.5\textpm0.9} & \textbf{74.4\textpm0.7} & \textbf{74.8\textpm0.8} & \textbf{44.0\textpm1.0} & \textbf{91.0\textpm0.5}   \\
    \end{tabular}
    \vspace{-0.1in}
    \caption{Cross-validated classification accuracy results. Note that * denotes that this dataset was excluded from training, and therefore, signifies out-of-domain performance. Values in bold indicate significant statistical gains over CNAPS.}
    \vspace{-0.1in}
    \label{class-validation-fine-grained-results}
\end{table*}{}

\begin{table*}
    \small
    \centering
    \begin{tabular}{l|ccc|ccc}
        & \multicolumn{3}{c}{Average Accuracy with $\psi_\phi^f$ (\%)} & \multicolumn{3}{c}{Average Accuracy without $\psi_\phi^f$ (\%)}\\
        Metric/Model Variant & In-Domain & Out-Domain & Overall & In-Domain & Out-Domain & Overall \\
        \hline
        Negative Dot Product & 66.9\textpm0.9 & 53.9\textpm0.8 & 61.9\textpm0.9 & 38.4\textpm1.0 & 44.7\textpm1.0 & 40.8\textpm1.0 \\
        CNAPS & 69.6\textpm0.8 & 59.8\textpm0.8 & 65.9\textpm0.8  & \textbf{54.4\textpm1.0} & 55.7\textpm0.9 & 54.9\textpm0.9 \\
        Absolute Distance ($L_1$) & 71.0\textpm0.8 & 65.4\textpm0.8 & 68.8\textpm0.8 & \textbf{54.9\textpm1.0} & 62.2\textpm0.8 & 57.7\textpm0.9  \\
        Squared Euclidean ($L_2^2$) & 71.7\textpm0.8 & 66.3\textpm0.8 & 69.6\textpm0.8 & \textbf{55.3\textpm1.0} & 61.8\textpm0.8 & \textbf{57.8\textpm0.9} \\
        \hline
        Simple CNAPS -TR & \underline{\textbf{73.5\textpm0.8}} & \underline{\textbf{69.6\textpm0.8}} & \underline{\textbf{72.0\textpm0.8}} & 52.3\textpm1.0 & \underline{61.7\textpm0.9} & 55.9\textpm1.0  \\
        Simple CNAPS & \underline{\textbf{73.8\textpm0.8}} & \underline{\textbf{69.7\textpm0.8}} & \underline{\textbf{72.2\textpm0.8}} & \textbf{56.0\textpm1.0} & \underline{\textbf{64.8\textpm0.8}} & \underline{\textbf{59.3\textpm0.9}}  \\
    \end{tabular}
    \vspace{-0.12in}
    \caption{Comparing in-domain, out-of-domain and overall accuracy averages of each metric/model variant when feature extractor adaptation is performed (denoted as "with $\psi_\phi^f$") vs. when no adaptation is performed (denoted as "without $\psi_\phi^f$"). Values in bold signify best performance in the column while underlined values signify superior performance of Simple CNAPS (and the -TR variant) compared to the CNAPS baseline.}
    \vspace{-0.1in}
    \label{feature-extractor-adaptation-ablation}
\end{table*}{}

\begin{table}[t]
    \small
    \centering
    \begin{tabular}{cc|cccc}
        & & \multicolumn{3}{c}{Average Classification Accuracy (\%)}\\
        Fold & Model & In-Domain & Out-Domain & Overall \\
        \hline
        1 & CNAPS & 70.1\textpm0.4& 51.6\textpm0.4& 65.5\textpm0.4   \\
        1 & S. CNAPS & \textbf{75.7\textpm0.3}& \textbf{64.0\textpm0.4}& \textbf{72.7\textpm0.3}   \\
        \hline
        2 & CNAPS & 69.3\textpm0.4& 48.9\textpm0.3& 64.2\textpm0.4  \\
        2 & S. CNAPS & \textbf{74.3\textpm0.4}& \textbf{69.3\textpm0.4} & \textbf{73.0\textpm0.3}    \\
        \hline
        3 & CNAPS & 68.4\textpm0.4& 48.5\textpm0.4& 63.4\textpm0.4   \\
        3 & S. CNAPS & \textbf{71.8\textpm0.4}& \textbf{67.3\textpm0.5}& \textbf{70.7\textpm0.4}    \\
        \hline
        4 & CNAPS & 67.9\textpm0.3& 51.0\textpm0.7& 63.7\textpm0.4   \\
        4 & S. CNAPS & \textbf{73.6\textpm0.3}& \textbf{62.6\textpm0.6}& \textbf{70.9\textpm0.4}   \\
        \hline
        Avg & CNAPS & 69.0\textpm1.4& 50.0\textpm1.8& 64.2\textpm1.6   \\
        Avg & S. CNAPS & \textbf{73.8\textpm1.3}& \textbf{65.8\textpm1.8}& \textbf{71.8\textpm1.4}   \\
    \end{tabular}
    \vspace{-0.1in}
    \caption{Cross-validated in-domain, out-of-domain and overall classification accuracies averaged across each fold and combined. Note that for conciseness, Simple CNAPS has been shortened to "S. CNAPS". Simple CNAPS values in bold indicate statistically significant gains over CNAPS.}
    \vspace{-0.1in}
    \label{class-validation-averages}
\end{table}{}

We adapt the same architectural choices for the task encoder $g_\phi$, auto-regressive set encoders $g_\phi^{{AZ}_1},...,g_\phi^{{AZ}_J}$ and the feature extractor adaptation network $\psi_\phi^f = \{\psi_\phi^1,...,\psi_\phi^J\}$ as \cite{requeima2019fast}. The neural architecture for each adaptation module inside of $\psi_\phi^f$ has been shown in Figure \ref{adaptation-network-architecture}. The neural configurations for the task encoder $g_\phi$ and the auto-regressive set encoders $g_\phi^{{AZ}_1},...,g_\phi^{{AZ}_J}$ used in AR-CNAPS are shown in Figure \ref{set-encoders-and-classifier}-a and Figure \ref{set-encoders-and-classifier}-b respectively. Note that for the auto-regressive set encoders, there is no need for convolutional layers. The input to these networks come from the output of the corresponding residual block adapted to that level (denoted by $f_\theta^{\tau_j}$ for block $j$) which has already been processed with convolutional filters.

Unlike CNAPS, we do not use the classifier adaptation network $\psi_\phi^c$. As shown in Figure \ref{set-encoders-and-classifier}-c, the classification weights adaptor $\psi_\phi^c$ consists of an MLP consisting of three fully connected (FC) layers with the intermediary none-linearity ELU, which is the continuous approximation to ReLU as defined below:

\begin{equation}
    \begin{split}ELU(x) = \begin{Bmatrix} x & x > 0 \\
    e^x – 1 & x \le 0 \end{Bmatrix}\end{split}
\end{equation}

As mentioned previously, without the need to learn the three FC layers in $\psi_\phi^c$, Simple CNAPS has 788,485 fewer parameters while outperforming CNAPS by considerable margins.

\section{Cross Validation}
\label{appendix:cross-validation}

The Meta-Dataset \cite{triantafillou2019meta} and its 8 in-domain 2 out-of-domain split is a setting that has defined the benchmark for the baseline results provided. The splits, between the datasets, were intended to capture an extensive set of visual domains for evaluating the models.

However, despite the fact that all past work directly rely on the provided set up, we go further by verifying that our model is not overfitting to the proposed splits and is able to consistently outperform the baseline with different permutations of the datasets. We examine this through a 4-fold cross validation of Simple CNAPS and CNAPS on the following 8 datasets: ILSVRC-2012 (ImageNet) \cite{russakovsky2015imagenet}, Omniglot \cite{lake2015human}, FGVC-Aircraft \cite{maji2013fine}, CUB-200-2011 (Birds) \cite{wah2011caltech}, Describable Textures (DTD) \cite{cimpoi2014describing}, QuickDraw \cite{jongejan2016quick}, FGVCx Fungi \cite{fungi2018schroeder} and VGG Flower \cite{nilsback2008automated}. During each fold, two of the datasets are exluded from training, and both Simple CNAPS and CNAPS are trained and evaluated in that setting. 

As shown by the classification results in Table \ref{class-validation-fine-grained-results}, in all four folds of validation, Simple CNAPS is able to outperform CNAPS on 7-8 out of the 8 datasets. The in-domain, out-of-domain, and overall averages for each fold noted in Table \ref{class-validation-averages} also show Simple CNAPS's accuracy gains over CNAPS with substantial margins. In fact, the fewer number of in-domain datasets in the cross-validation (6  vs. 8) actually leads to wider gaps between Simple CNAPS and CNAPS. 
This suggests Simple CNAPS is a more powerful alternative in the low domain setting.
Furthermore, using these results, we illustrate that our gains are not specific to the Meta-Dataset setup.

\section{Ablation study of the Feature Extractor Adaptation Network}

In addition to the choice of metric ablation study referenced in Section \ref{sec:experiments-ablation}, we examine the behaviour of the model when the feature extractor adaptation network $\psi_\phi^f$ has been turned off. In such setting, the feature extractor would only consist of the pre-trained ResNet18 \cite{DBLP:journals/corr/HeZRS15-resnet} $f_\theta$. Consistent to \cite{requeima2019fast}, we refer to this setting as "No Adaptation" (or ``No Adapt" for short). We compare the ``No Adapt" variant to the feature extractor adaptive case for each of the metrics/model variants examined in Section \ref{sec:experiments-ablation}. The in-domain, out-of-domain and overall classification accuracies are shown in Table \ref{feature-extractor-adaptation-ablation}. As shown, without $\psi_\phi^f$ all models lose approximately 15, 5, and 12 percentage points across in-domain, out-of-domain and overall accuracy, while Simple CNAPS continues to hold the lead especially in out-of-domain classification accuracy. It's interesting to note that without the task specific regularization term (denoted as "-TR"), there's a considerable performance drop in the ``No Adaptation" setting; while when the feature extractor adaptation network $\psi_\phi^f$ is present, the difference is marginal. This signifies two important observations. First, it shows the importance of of learning the feature extractor adaptation module end-to-end with the Mahalanobis distance, as it's able adapt the feature space best suited for using the squared Mahalanobis distance. Second, the adaptation function $\psi_\phi^f$ can reduce the importance of the task regularizer by properly de-correlating and normalizing variance within the feature vectors. However, where this is not possible, as in the ``No Adaptation" case, the all-classes-task-level covariance estimate as an added regularizer in Equation \ref{mh-distance} becomes crucial in maintaining superior performance.

\section{Projection Networks}
We additionally explored metric learning where in addition to changing the distance metric, we considered projecting each support feature vector $f_\theta^\tau(\vx_i)$ and query vector $f_\theta^\tau(\vx_i^*)$ to a new decision space where then squared Mahalanobis distance was to be used for classification. 
Specifically, we trained a projection network $u_\phi$ such that for Equations \ref{mh-distance} and \ref{cov-regularized-task-cov}, $\vmu_k$, $\mathbf{\Sigma}_k^\tau$ and $\mathbf{\Sigma}^\tau$ were calculated based on the projected feature vectors $\{u_\phi(f_\theta^\tau(\vx_i))\}_{\vx_i \in \mathcal{S}^\tau_k}$ as oppose to the feature vector set $\{f_\theta^\tau(\vx_i)\}_{\vx_i \in \mathcal{S}^\tau_k}$. Similarly, the projected query feature vector $u_\phi(f_\theta^\tau(\vx_i^*))$ was used for classifying the query example as oppose to the bare feature vector $f_\theta^\tau(\vx_i^*)$ used within Simple CNAPS. We define $u_\phi$ in our experiments to be the following:
\begin{equation}
    u_\phi(f_\theta^\tau(\vx_i^*)) = W_1(ELU(W_2(ELU(W_3f_\theta^\tau(\vx_i^*)))))
\end{equation}
where ELU, a continuous approximation to ReLU as previously noted, is used as the choice of non-linearity and $W_1$, $W_2$ and $W_3$ are learned parameters. 

\begin{table}[t]
    \small
    \centering
    \begin{tabular}{l|ccc}
         & \multicolumn{3}{c}{Average Classification Accuracy (\%)} \\
         Model & In-Domain & Out-Domain & Overall \\
         \hline
         Simple CNAPS +P & 72.4\textpm0.9 & 67.1\textpm0.8 & 70.4\textpm0.8 \\
         Simple CNAPS & 73.8\textpm0.8 & \textbf{69.7\textpm0.8} & 72.2\textpm0.8
    \end{tabular}
    \vspace{-0.1in}
    \caption{Comparing the in-domain, out-of-domain and overall classification accuracy of Simple CNAPS +P (with projection networks) to Simple CNAPS. Values in bold show the statistically significant best result.}
    \vspace{-0.1in}
    \label{projection-network-results}
\end{table}

We refer to this variant of our model as ``Simple CNAPS +P" with the ``+P" tag signifying the addition of the projection function $u_\phi$. The results for this variant of Simple CNAPS are compared to the base Simple CNAPS in Table \ref{projection-network-results}. As shown, the projection network generally results in lower performance, although not to statistically significant degrees in in-domain and overall accuracies. Where the addition of the projection network results in substantial loss of performance is in the out-of-domain setting with Simple CNAPS +P's average accuracy of 67.1\textpm0.8 compared to 69.7\textpm0.8 for the Simple CNAPS. We hypothesize the significant loss in out-of-domain performance to be due to the projection network overfitting to the in-domain datasets.

\end{appendices}

\end{document}